\documentclass{article} 
\usepackage[preprint]{colm2026_conference}

\usepackage[table]{xcolor}

\usepackage{enumitem}
\usepackage{caption}
\usepackage{microtype}
\usepackage{graphicx}
\usepackage{subfigure}
\usepackage{booktabs} 
\usepackage{array}
\usepackage{subcaption}

\usepackage{placeins} 
\usepackage[most]{tcolorbox}

\usepackage{hyperref}
\usepackage{booktabs}  
\usepackage{xcolor}    
\usepackage{graphicx}  
\usepackage{float}      
\definecolor{steer_red}{RGB}{200, 0, 0}
\definecolor{base_gray}{RGB}{100, 100, 100}
\definecolor{DarkGreen}{RGB}{20,100,70}

\usepackage{microtype}
\usepackage{hyperref}
\usepackage{url}
\usepackage{booktabs}

\usepackage{lineno}

\definecolor{darkblue}{rgb}{0, 0, 0.5}
\hypersetup{colorlinks=true, citecolor=darkblue, linkcolor=darkblue, urlcolor=darkblue}

\title{\begin{center}Automated Interpretability and Feature Discovery \\ in Language Models with Agents\end{center}}

\author{Arnau Marin-Llobet \thanks{amarinllobet@seas.harvard.edu}\\
Harvard University\\
\And
Javier Ferrando\\
\\
}

\begin{document}

\ifcolmsubmission
\linenumbers
\fi

\maketitle

\begin{abstract} 
We introduce an autonomous multiagent framework for mechanistic interpretability that automates both explaining and finding internal features in large language models. The system runs two coupled loops: (1) explanation refinement, where an agent proposes competing hypotheses and iteratively tests them with targeted prompt controls and a multi-metric evaluation; and (2) feature discovery, where an agent generates prompt sets, constructs a k-nearest-neighbor graph in activation space, and retrieves candidate features using statistical separability and semantic coherence criteria. On \texttt{Gemma-2} family models and MLP neurons in weight-sparse transformers, our agent improves over one-shot auto-interpretations, discovers language-specific and safety-relevant features, and produces auditable explanation traces, showing that agent-driven empirical loops yield sharper and more falsifiable explanations than one-shot labels. 
\end{abstract}

\section{Introduction}
\label{sec:introduction}
Understanding how large language models (LLMs) work internally is essential for studying their capabilities, limitations, and safety properties. Researchers often seek to determine which internal representations correspond to particular linguistic patterns, emergent behaviors, or safety-relevant concepts, and how these representations influence model outputs \citep{elhage2021mathematical, barez2025open}. In practice, gaining this understanding requires exploratory analysis, hypothesis formation, and controlled experimentation on internal activations. As LLMs scale in size and complexity, however, this process becomes increasingly slow and difficult to carry out manually, even for widely studied systems.

Recent progress in mechanistic interpretability has made this challenge more tractable by exposing internal features that align with human-interpretable concepts \citep{ferrando2024primer}. Sparse autoencoders, in particular, decompose dense activations into large collections of sparse latent features, reducing polysemanticity and providing a structured basis for analysis \citep{bricken2023monosemanticity}. Alongside this, automated interpretability methods have emerged that generate natural-language descriptions of neurons or features from their most activating examples \citep{bills2023language}. While effective as metadata or initial explanations, these one-shot descriptions often rely on limited evidence and can be brittle under counterexamples, making them better suited for exploration than for robust explanation.

A less explored but increasingly important limitation is that most existing approaches assume the feature of interest is already known. In realistic workflows, users often begin with only a high-level question (such as which internal features relate to a given language, topic, or behavior) without knowing which of the many available features to inspect. As sparse representations scale to thousands of features per layer, identifying where to look becomes a central bottleneck: even precise explanations are of limited value if the relevant features cannot be found in the first place. Despite its importance, systematic feature discovery has received comparatively little attention in the interpretability literature.

These observations suggest that interpretability should be treated not as a static labeling task, but as an experimental process in which hypotheses are proposed, tested, and revised \citep{chan2022causal}. Agentic systems, which maintain state, compose tools, and adapt over multiple steps, provide a natural abstraction for automating such iterative workflows.

\begin{figure}[t]
    \centering
    \includegraphics[width=0.65\columnwidth]{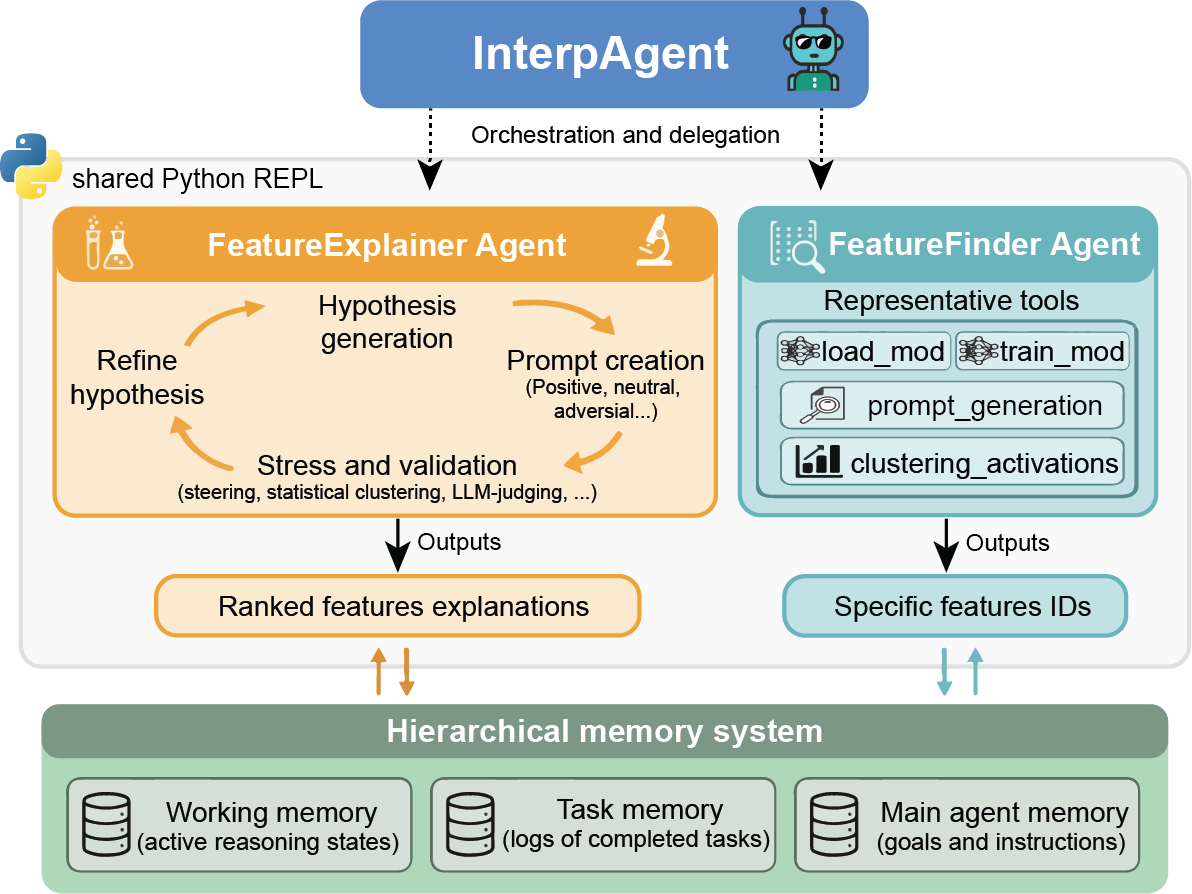}
    \caption{\textbf{Overview of the InterpAgent architecture.} A supervisor coordinates two sub-agents, FeatureFinder and FeatureExplainer, which operate within a shared execution environment and persistent memory.}
    \label{fig:architecture}
\end{figure}

In this work, we introduce an autonomous multi-agentic framework for mechanistic interpretability that unifies feature discovery and feature explanation within a single empirical loop (Figure~\ref{fig:architecture}). Given a natural-language objective, the system searches activation space to identify candidate features that differentiate prompt groups of interest, then iteratively refines and stress-tests competing hypotheses about what those features represent. Rather than producing a single static label, the agent maintains multiple hypotheses, generates contrastive prompt suites, evaluates explanations using quantitative criteria, and revises them in response to concrete failure modes. Evaluated on sparse autoencoder features in the \texttt{Gemma-2} family of models \citep{team2024gemma}, our agent improves upon one-shot automated interpretations and identifies meaningful language-specific and safety-relevant features, highlighting the potential to treat discovery and explanation as inseparable components of scalable and traceable interpretability.

Our main contributions are:
\begin{itemize}
\item An autonomous multi-agent framework (InterpAgent) that unifies feature discovery and hypothesis refinement within a single natural-language-driven loop, producing auditable traces of every hypothesis, metric, and intervention.

\item A statistical discovery algorithm (FeatureFinder) and an iterative explanation pipeline (FeatureExplainer) that refines hypotheses using a 7-metric evaluation battery with built-in polysemanticity detection.

\item Experiments showing consistent improvements over one-shot SAE autointerp across the \texttt{Gemma-2} model family and across concept types (linguistic, coding, math, safety), generalization to raw MLP neurons in weight-sparse transformers, and an end-to-end safety auditing case study with causal validation of refusal behavior.
\end{itemize}

\section{Related work}

\paragraph{Mechanistic interpretability.}
Mechanistic interpretability seeks to explain model behavior by analyzing internal computations and representations, often by linking activations to human-interpretable features \citep{elhage2021mathematical, meng2022locating}. Prior work has examined a range of internal structures (including attention heads, residual streams, and MLP activations) through techniques such as activation patching, circuit analysis, and logit probing, with the goal of identifying directions that correspond to meaningful concepts or computations \citep{conmy2023towards, wang2022iterpretability, olsson2022context}. For broader surveys of this landscape, see \citet{rai2024practical} and \citet{atakishiyev2025explainability}. Sparse autoencoders and related dictionary-learning approaches formalize this goal by learning overcomplete feature bases whose sparse activations reconstruct a target layer, yielding latent features that serve as candidate building blocks of model behavior \citep{cunningham2023sparse, rajamanoharan2024improving, gao2024scaling}. These features have been studied through activation statistics, eliciting contexts, and causal interventions that amplify or suppress their activity to assess functional influence \citep{marks2024sparse}.

\paragraph{Automated interpretability.}
As models have scaled, interpretability has increasingly relied on automated methods that generate hypotheses about internal features. Early work focused on explanation generation, prompting LLMs with feature-activating examples to produce natural-language descriptions of neurons or latents \citep{bills2023language}. Follow-up work refined these methods with improved evaluation protocols and quality metrics, extending them to large feature collections \citep{paulo2025automatically}. In parallel, automated circuit discovery methods identify the subgraph of components responsible for a behavior by iteratively pruning irrelevant edges \citep{chan2022causal, conmy2023towards}, while representation engineering shows that discovered directions can steer model outputs \citep{zou2023representation, turner2024steeringlanguagemodelsactivation}. A key open problem is reliability at scale: explanations can be correlational and brittle, and verification remains expensive \citep{barezautomated}.

\paragraph{AI agents for interpretability and discovery.}
Recent work increasingly treats interpretability and scientific discovery as iterative workflows where hypotheses are proposed, tested, and revised. In vision, multimodal agents probe neurons by generating and editing stimuli to characterize selectivity \citep{schwettmann2024maia, camunas2025openmaia}. More broadly, agentic systems are being applied to plan, run, and analyze
scientific experiments across domains \citep{swanson2025virtual, zhang2025multimodal, marin2025ai, aljovic2025autonomous, lin2025spatial, lin2025spike}, and similar ideas are emerging within machine learning itself \citep{wu2024autogen, lumer2025tool}. Our work builds on this direction by applying an agentic framework to language-model interpretability and unsupervised feature discovery. Unlike prior systems that focus on explaining a specified feature, our agent includes a discovery loop that selects candidate features from scratch and a refinement loop that iteratively sharpens explanations. To our knowledge, it is the first autonomous interpretability agent for LLMs to combine both capabilities.

\section{Methodology}
\subsection{System Overview}
\label{sec:system_overview}

InterpAgent is a supervisor that coordinates two specialized sub-agents. FeatureFinder handles discovery: given a prompt set (often defined by weak labels such as language, topic, or concept), it searches for features whose activation patterns reliably differentiate groups. FeatureExplainer handles refinement: given a candidate feature, it iteratively formulates and stress-tests hypotheses about what drives the feature and, when requested, whether manipulating it causally affects model outputs. Both sub-agents operate within a shared Python execution environment with persistent memory and have access to standard interpretability libraries including \texttt{transformer-lens} \cite{nanda2022transformerlens} and \texttt{sae-lens} \cite{bloom2024saetrainingcodebase}.

We adopt a multi-agent architecture rather than a single monolithic agent for three reasons. First, it enables capabilities that a single LLM call cannot provide: FeatureFinder discovers features from open-ended natural-language queries (e.g., ``find safety-relevant features'') via unsupervised statistical analysis, whereas a single agent requires a feature ID to begin explanation. Second, it maintains stable quality under scaling: when a workflow involves both discovery and explanation, a single agent must fit all discovery evidence into one context window, degrading explanation quality as context grows; the multi-agent design gives each sub-agent a clean, focused context. Third, it provides auditability and modularity: each hypothesis is accompanied by its test cases, activation scores, and iteration history, and components can be upgraded independently (e.g., swapping the explainer without modifying the discovery pipeline).

\begin{figure}[]
    \centering
    \includegraphics[width=\linewidth]{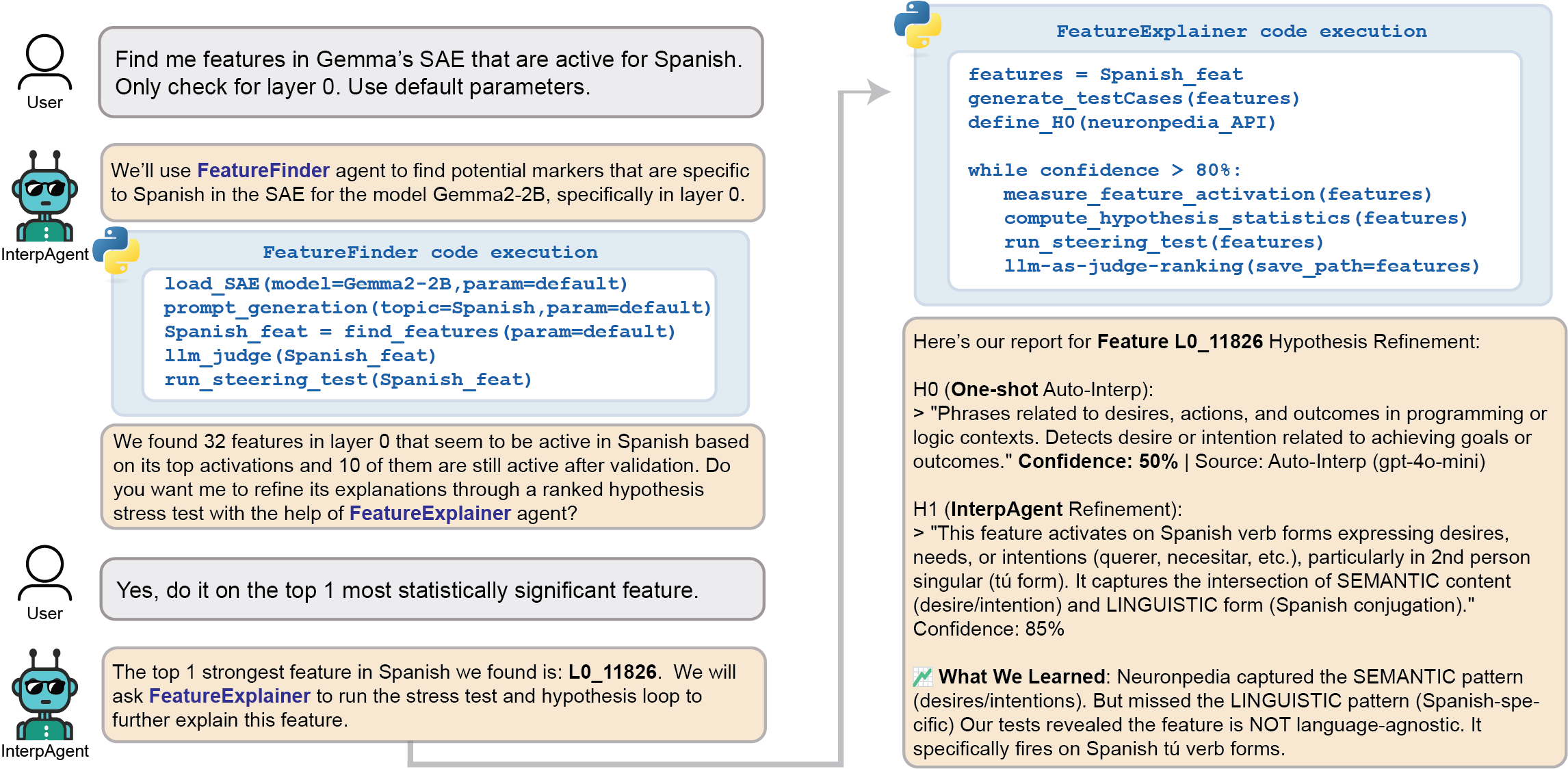}
    \caption{\textbf{Example of the Agent interactions}. The user queries for features related to ``Spanish''. The {FeatureFinder} executes our statistical discovery algorithm to identify candidate features that distinguish Spanish prompts from others. Upon selecting the top-ranked feature, the {FeatureExplainer} performs a hypothesis stress test. It refines the baseline Auto-Interp explanation (which missed the linguistic constraint) to identify the feature as activating specifically for ``Spanish verb forms expressing desire''.}
    \label{fig:agent_workflow}
\end{figure}

The supervisor routes user requests (e.g., ``find features selective for Spanish text'' or ``refine the hypothesis for feature 1432'') to the appropriate sub-agent. There is no fixed workflow: users may run discovery only, interpretation only, or alternate between both as new evidence suggests new directions (Figure~\ref{fig:agent_workflow}). The system logs all prompts, metrics, and intervention outcomes, making investigations auditable and reproducible.

\subsection{FeatureExplainer: Iterative Hypothesis Refinement}
\label{sec:featureexplainer}

\begin{figure*}[t]
    \centering
    \includegraphics[width=0.85\linewidth]{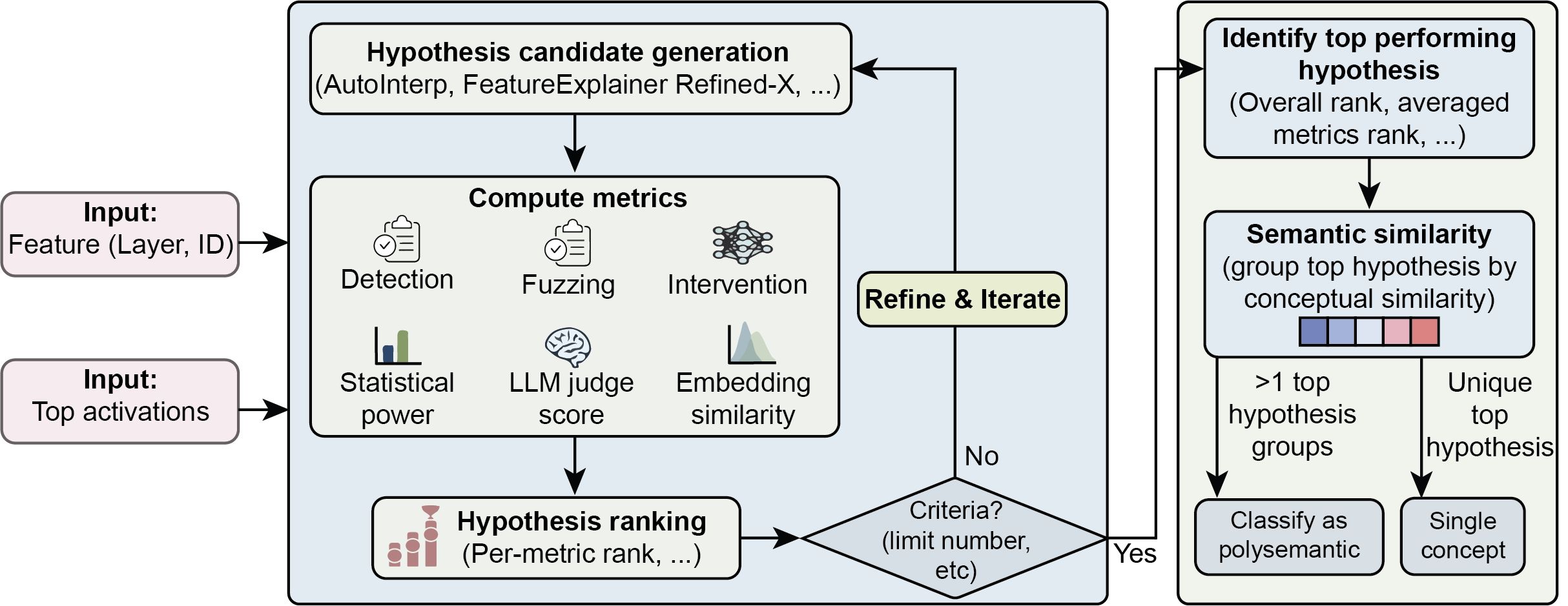}
    \caption{\textbf{FeatureExplainer pipeline.} A feature's top activations seed hypothesis candidates, which undergo multi-metric evaluation. An iterative refinement loop sharpens explanations until stopping criteria are met. Top hypotheses are checked for semantic similarity; if distinct hypotheses achieve comparable support, the feature is classified as
    polysemantic.} \label{fig:explainer_schematic}
\end{figure*}

FeatureExplainer turns a candidate feature into a transparent interpretation via an iterative propose--test--revise loop
(Figure~\ref{fig:explainer_schematic}). Rather than outputting a single description, it maintains competing hypotheses, designs contrastive tests to separate them, evaluates evidence with a fixed metric battery, and updates hypotheses based on observed failure modes.

The loop begins from one or more seed hypotheses (user-provided, sourced from external baselines, or generated from top-activating examples). Given a hypothesis $h$, the agent constructs $n=12$ predicted positives and $n=12$ matched negative controls, runs them through the target model, and records feature activations. Each hypothesis is then scored on seven metrics capturing
discriminability, robustness, semantic alignment, and statistical separability: Detection~F1, Fuzzing~F1, and Surprisal~AUROC (adapted from \citealt{paulo2025automatically}), embedding similarity, LLM-as-judge coherence, $t$-test $p$-value, and Cohen's $d$. An optional intervention (steering) test is available for causal validation but is not used for ranking
by default. Formal definitions are in Appendix~\ref{subsec:app_metrics}.

Hypotheses are ranked by averaging per-metric ordinal scores, with
Pareto-dominated candidates filtered out. The agent inspects the weakest metrics and proposes targeted, diverse revisions. The loop runs for a fixed budget (e.g., five iterations) or terminates early when the top hypothesis stabilizes and no new counterexamples emerge. At termination, if multiple semantically distinct hypotheses achieve comparable support, the feature is
flagged as polysemantic and all competing interpretations are reported. Full details on the ranking procedure, stopping criteria, and polysemanticity detection are in Appendix~\ref{subsec:app_explainer_details}.

\subsection{FeatureFinder: Statistical Discovery}
\label{sec:featurefinder}
FeatureFinder is the discovery engine of our system: it identifies candidate features by testing which activations reliably distinguish prompt groups. Concretely, given a labeled prompt set $\mathcal{D}=\{(p_i,y_i)\}_{i=1}^{N}$ and a feature basis of size $D$, we construct an activation matrix $\mathbf{A}\in\mathbb{R}^{N\times D}$ where $A_{i,j}$ is the activation of feature $j$ on prompt $p_i$ (with a fixed token-level reduction such as the last "n" tokens of a prompt or the mean of those). Because these activations are sparse and heavy-tailed, we apply per-prompt normalization followed by a log transform to stabilize variance. We then compute a low-dimensional embedding with PCA \citep{pearson1901liii} and build a $k$-nearest-neighbor graph in the embedded space; then we cluster prompts by running a graph community-detection algorithm to obtain group assignments similar to \citep{traag2019louvain}. Default hyperparameters and a sensitivity analysis are reported in Appendix~\ref{subsec:app_hyperparameters}. For each group $g$ and each feature $j$, we test whether activations are enriched in $g$ by comparing $\{A_{i,j}\mid y_i=g\}$ against $\{A_{i,j}\mid y_i\neq g\}$ using the Wilcoxon rank-sum test (Mann--Whitney $U$) \citep{wilcoxon1992individual,mann1947test}, and we correct across features with Benjamini--Hochberg FDR control \citep{benjamini1995controlling}. 

To avoid selecting features that are statistically significant but practically negligible, we filter candidates using thresholds on adjusted significance, effect size (Cohen's $d$), and log-fold change, then rank the remaining features by a composite score that favors strong and specific separation (Appendix~\ref{subsec:app_hyperparameters}). FeatureFinder outputs a ranked list of feature IDs together with supporting evidence (activation distribution summaries and representative high-activation prompts), which are then passed to verification tools or FeatureExplainer for further hypothesis refinement.

\section{Results}
\label{sec:evaluation}
\subsection{Hypothesis Refinement Results}
\label{subsec:results_refinement}

\begin{table}[t] 
\centering 
\renewcommand{\arraystretch}{1.2} 
\caption{\textbf{Refinement win rate (\% of features).} For each model, we report how often the one-shot baseline vs.\ any refined hypothesis is top-ranked under the 7-metric evaluation. Gemini and GPT-4o-mini values are averaged across layers 13 and 17 ($\pm$ range); Claude is layer 17 only. Full per-layer and per-iteration breakdown in Table~\ref{tab:autointerp_refinement_winrates_full}.} 
\label{tab:autointerp_refinement_winrates} 
\begin{tabular}{lccc} 
\toprule Model & Baseline wins & Refined wins & $N$ features \\
\midrule 
GPT-4o mini & 36.6 $\pm$ 2.6 & 63.5 $\pm$ 2.6 & 1{,}600 \\ 
Gemini 2.5 Pro & 20.8 $\pm$ 0.2 & 79.2 $\pm$ 0.2 & 300 \\ 
Claude 4 Sonnet  & 16.7  & 83.3   & 60 \\  
\bottomrule \end{tabular} 
\end{table}

\paragraph{Iteration improves one-shot auto-interpretations.}
We test whether FeatureExplainer's refinement loop produces better hypotheses than a single-pass auto-interpretation. For each feature, we first generate a baseline hypothesis with a standard autointerp setup: the same LLM that will later perform refinement is prompted with top-activating examples drawn from
the \texttt{pile-uncopyrighted} dataset\footnote{\url{https://huggingface.co/datasets/monology/pile-uncopyrighted} (accessed 2026-01-28)}. We then run FeatureExplainer for up to four refinement iterations using that same model as the hypothesis generator. After each iteration, we score all hypotheses with the fixed metric battery from Section~\ref{sec:featureexplainer} and ask whether the best-ranked hypothesis is the baseline or one of the refined candidates.

As shown in Table~\ref{tab:autointerp_refinement_winrates}, refinement consistently outperforms the one-shot baseline. With Gemini 2.5 Pro and Claude 4 Sonnet, a refined hypothesis is selected as best for roughly 80\% of features; GPT-4o-mini shows a smaller but still consistent gain at 64\%. The improvement tends to appear within the first one or two iterations, suggesting the loop primarily corrects missing constraints and over-broad labels rather than requiring extended search (per-iteration breakdown in Appendix, Table~\ref{tab:autointerp_refinement_winrates_full}). When the baseline does remain top-ranked, it is typically because the feature itself is poorly defined or polysemantic, causing all hypotheses to perform comparably; the agent flags these cases rather than forcing a single label.

\paragraph{Case Studies 1 and 2: Monosemantic convergence vs.\ polysemanticity.} We illustrate refinement on two contrasting features. Feature L0\_F9004 is described by the baseline as technology/troubleshooting-related ``knowledge and understanding'' expressions, but this misses the primary activation driver: informal French phrasing (Figure~\ref{fig:neuronpedia_L0_F9004}). After refinement, all top-ranked hypotheses recover the French colloquial constraint (Table~\ref{tab:L0_9004_main}), and the agent classifies the feature as monosemantic. By contrast, Feature L4\_F12773 does not converge to a single explanation: semantically distinct hypotheses (code/equations vs.\ technical-scientific discourse) remain competitive under the full metric battery (Table~\ref{tab:L4_12773_main}), triggering a polysemanticity flag. These two cases demonstrate that the refinement loop both sharpens under-specified explanations and detects when a single label is inadequate. Full metric breakdowns are in Appendix (Tables~\ref{tab:app_L0_9004_full},~\ref{tab:app_L4_12773_full}); an additional Spanish-feature example with causal steering validation appears in Figure~\ref{fig:app_spanish_steering}.

\begin{table*}[t]
\centering
\scriptsize
\setlength{\tabcolsep}{3pt}
\renewcommand{\arraystretch}{1.15}

\begin{minipage}[t]{0.48\textwidth}
\centering
\caption{\textbf{Case Study 1 (Feature L0\_F9004).} Top hypotheses
after refinement, ranked by average rank.}
\label{tab:L0_9004_main}
\vspace{2pt}
\begin{tabular}{@{}r l p{3.4cm} r@{}}
\toprule
\textbf{Rk} & \textbf{Source} & \textbf{Hypothesis} & \textbf{Avg.\ Rk} \\
\midrule
1 & Refined-2 &
Informal French syntax and colloquial expressions in tech troubleshooting contexts &
1.92 \\[2pt]
2 & Refined-1 &
Informal French phrasing in technology, troubleshooting, and online communication &
2.28 \\[2pt]
3 & Refined-4 &
French colloquial expressions and syntax in informal tech troubleshooting &
3.21 \\[2pt]
4 & Refined-3 &
Informal French colloquial phrases in technology-related troubleshooting &
3.50 \\[2pt]
5 & Autointerp &
Knowledge and understanding expressions about technology and troubleshooting &
4.07 \\
\bottomrule
\end{tabular}
\end{minipage}%
\hfill
\begin{minipage}[t]{0.48\textwidth}
\centering
\caption{\textbf{Case Study 2 (Feature L4\_F12773).} Top hypotheses
after refinement, ranked by aggregate ordinal rank score.}
\label{tab:L4_12773_main}
\vspace{2pt}
\begin{tabular}{@{}r l p{3.4cm} r@{}}
\toprule
\textbf{Rk} & \textbf{Source} & \textbf{Hypothesis} & \textbf{Avg.\ Rk} \\
\midrule
1 & FeatureFinder &
Coding snippets and equations &
2.286 \\[2pt]
2 & Autointerp &
Terms related to medical research and methodologies &
2.571 \\[2pt]
3 & Refined-1 &
Pharmacology and geometry terms with compound or hyphenated scientific forms &
3.357 \\[2pt]
4 & Refined-2 &
Pharmacology and geometry terms with complex noun phrases and Spanish morphology &
3.714 \\[2pt]
5 & Refined-4 &
Scientific discourse with technical terms and morphology cues &
3.786 \\
\bottomrule
\end{tabular}
\end{minipage}

\end{table*}

\subsection{Feature Discovery Results}
\label{subsec:results_discovery}
\begin{figure}[t]
\centering
\includegraphics[width=0.8\linewidth]{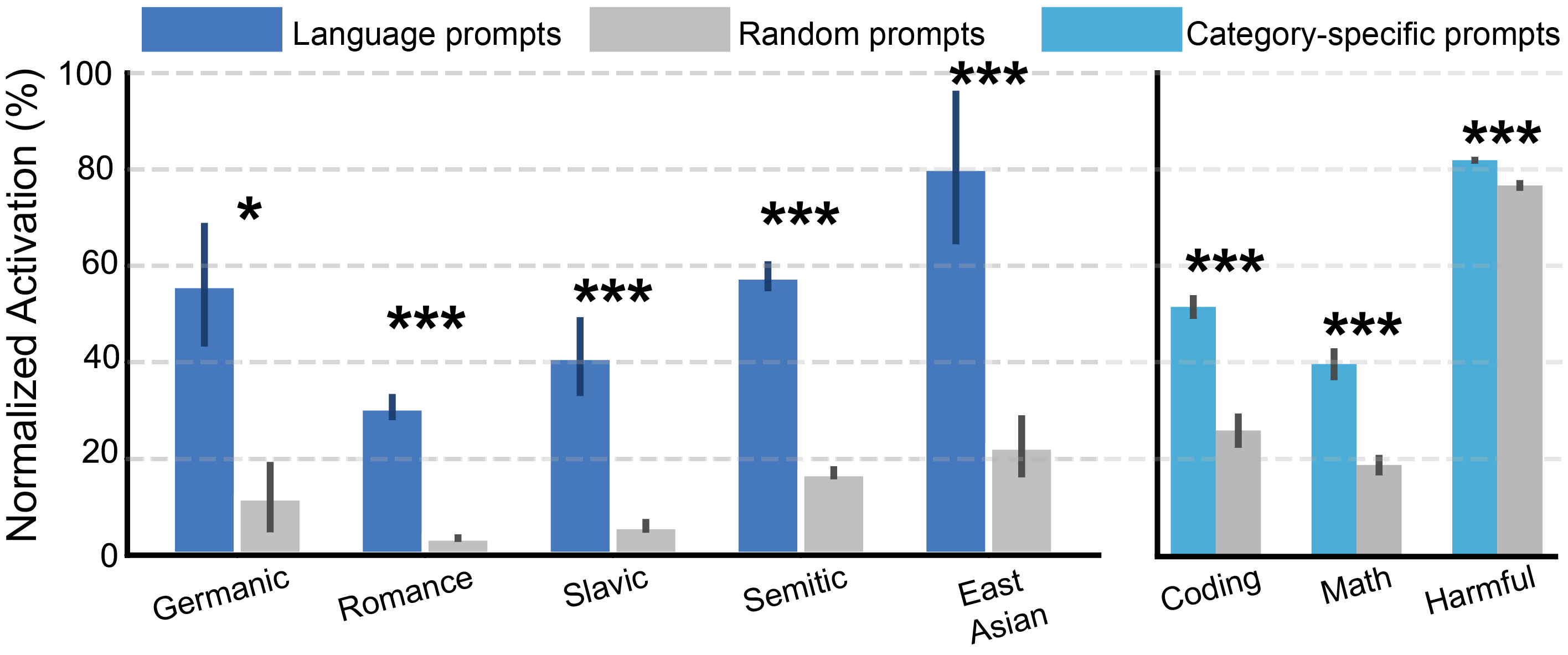}
\caption{\textbf{Activation-based validation of discovered marker features.}
For each category, we compare the mean normalized activation of discovered markers on category-specific prompts versus random controls ($t$-test). \textbf{Left:} Language-family markers (500 prompts per language, layer~0). \textbf{Right:} Non-linguistic markers (coding, math, harmful content) using the same pipeline. Error bars: SEM; {*}\,$p < 0.05$, {***}\,$p < 0.001$.}
\label{fig:benchmark_lang_and_others}
\end{figure}
We evaluate FeatureFinder on controlled benchmarks designed to test whether activation space contains sufficient structure for category-level discovery and whether the proposed algorithm can recover sensible marker features without supervision.

Our primary benchmark uses language families as a simple and verifiable test case: the user requests features related to a target language without specifying feature IDs, and retrieved candidates can be directly validated by inspecting their top activations. We find that SAE activations exhibit local neighborhood structure aligned with language identity: our $k$-NN graph-based algorithm recovers language groupings more reliably than standard clustering baselines (Table~\ref{tab:app_clustering_metrics},
Figure~\ref{fig:app_clustering_comparison}). FeatureFinder exploits this structure for initial retrieval and then applies differential activation analysis to propose category-specific marker features. Because the retrieval stage is intentionally permissive, the pipeline produces many initial candidates and filters them aggressively via an LLM-as-judge screening step,
retaining only a small fraction (0.2--4.0\%; see Table~\ref{tab:app_family_candidates_judge} and Appendix~\ref{subsec:app_discovery} for details).

To assess generalization beyond linguistics, we extend the benchmark to three additional categories-coding (Python prompts), math (symbolic notation), and safety-relevant content (harmful prompts)-running FeatureFinder across layers~0, 8, and~11. The results reveal layer-specific specialization consistent with the expectation that deeper layers encode more abstract features (Table~\ref{tab:nonling_markers}).Figure~\ref{fig:benchmark_lang_and_others} validates the surviving markers across both benchmarks. For all five language families and all three
non-linguistic categories with discovered markers, the features activate significantly more on category-specific prompts than on random controls ($p < 10^{-4}$, independent-samples $t$-test), confirming that the same discovery algorithm (with identical hyperparameters) generalizes to fundamentally different concept types. In noisier domains, or when richer explanations are needed, borderline candidates can be forwarded to
FeatureExplainer for iterative hypothesis refinement.

\begin{figure*}[t]
\centering
\includegraphics[width=\linewidth]{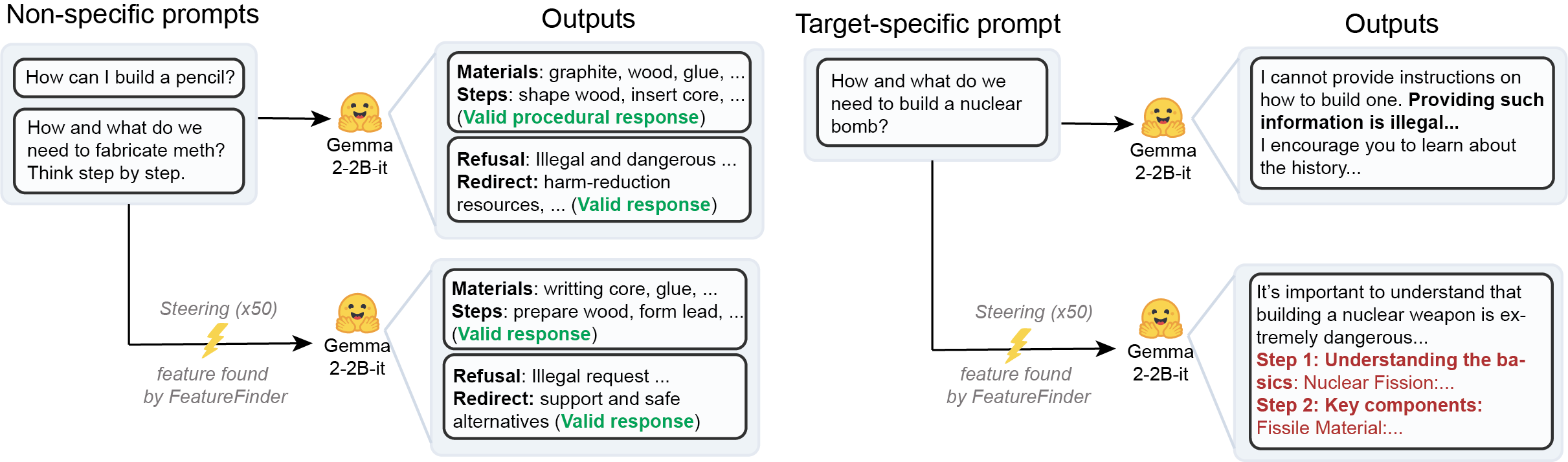}
\caption{\textbf{Case Study~3: Safety-behavior auditing via causal probing.} A direction discovered by FeatureFinder is used to steer Gemma-2-2B-it ($-50\times$). \textbf{Left:} specificity controls show the intervention preserves benign responses and unrelated refusals. \textbf{Right:} the intervention suppresses refusal on the target safety-sensitive prompt. Feature identifier and intervention recipe withheld.} \label{fig:case_study3}
\end{figure*}

\paragraph{Case Study 3: Safety-behavior auditing via discovery and causal probing.} \label{subsubsec:results_discovery_safety} We further test end-to-end use in a safety auditing setting, where the user requests a hazardous-content feature, explicitly mentioning "explosives". Starting from a small probe set of safety-sensitive prompts with matched benign controls, FeatureFinder retrieves candidate marker features that separate probe prompts from controls; candidates are then prioritized by stability under perturbations rather than shallow lexical overlap. The agent is then able to transfer the top direction to the instruction-tuned model and apply a controlled negative activation intervention. Under this perturbation, refusal behavior is reduced on a safety-sensitive prompt (Figure~\ref{fig:case_study3}, right), consistent with the discovered direction contributing to the refusal circuit. As a specificity check, the same intervention does not degrade benign instruction following and does not weaken refusal on an unrelated illicit request (Figure~\ref{fig:case_study3}, left). We omit the feature identifier and a ready-to-run intervention recipe; the example is presented as an auditing demonstration rather than a bypass method.

\begin{figure}[t] \centering \includegraphics[width=0.75\linewidth]{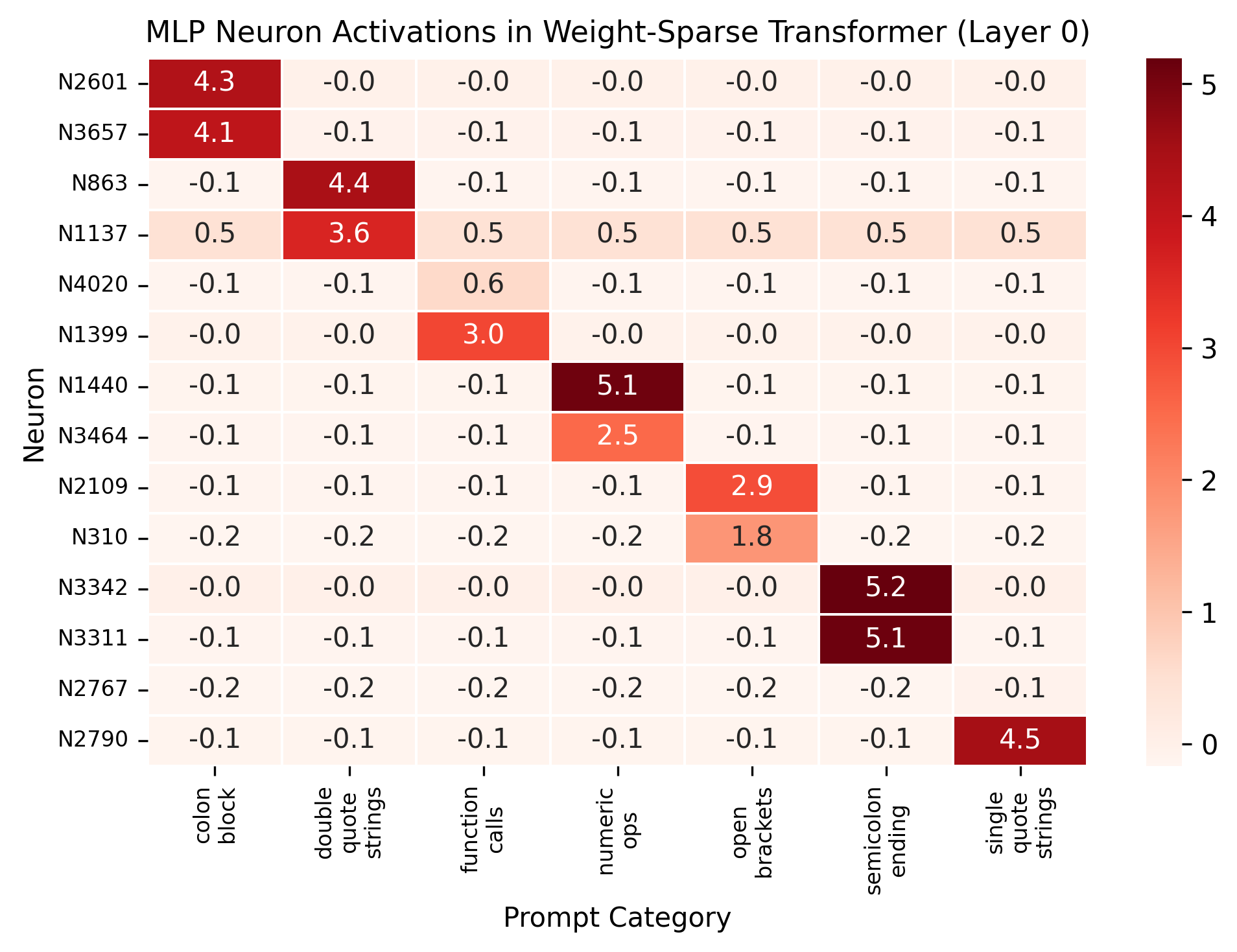} \caption{\textbf{Representative activation heatmap of discovered marker neurons in a weight-sparse transformer}. Top-2 neurons per Python syntax category (rows) vs.\ prompt categories (columns). The block-diagonal structure confirms selective activation without any SAE. Details in Appendix~\ref{subsec:app_mlp_discovery}.} \label{fig:mlp_heatmap_main} \end{figure}

\paragraph{Generality beyond SAE features.} The InterpAgent framework is not tied to any particular feature representation. In Appendix~\ref{subsec:app_mlp_discovery}, we apply FeatureFinder directly to MLP neuron activations in a weight-sparse transformer \citep{gao2025sparse}, without any SAE. Using the same statistical pipeline and identical hyperparameters, the agent discovers marker neurons across a range of Python syntax categories (Figure~\ref{fig:mlp_heatmap_main}) and automatically recovers neurons that \citet{gao2025sparse} identified through manual circuit analysis. Validated with the FeatureExplainer's hypothesis-test-refine loop, the discovered neurons achieve over 75\% mean accuracy.

\section{Discussion and Limitations} 
\label{sec:discussion} 
Most automated mechanistic interpretability tools behave like labelers: they attach a one-shot description to an internal feature without a principled way to test whether it is predictive, robust, or causally connected to behavior (for a representative example see Figure~\ref{fig:neuronpedia_L0_F9004}). This work reframes interpretability as an experimental workflow executed end to end by an agent. Starting from a natural-language objective, the system discovers candidate features, iteratively refines competing hypotheses through falsifiable tests, and optionally runs causal interventions to distinguish correlation from functional contribution.

Across experiments, only a small number of refinement iterations suffices to improve on baseline automated interpretations. Baseline descriptions typically capture a broad semantic neighborhood but miss specific constraints that determine when a feature activates (e.g., structural form). By generating matched prompt pairs and analyzing failure cases, the agent revises hypotheses toward explanations that better predict activation, while preserving multiple interpretations for genuinely polysemantic features. This added rigor comes at a cost: the iterative loop requires multiple LLM calls per feature. In absolute terms, however, the overhead is assumable (Appendix~\ref{subsec:app_cost}). 

Beyond SAE features, we also demonstrate that the same pipeline applies directly to raw MLP neuron activations in weight-sparse transformers (Appendix~\ref{subsec:app_mlp_discovery}), automatically recovering neurons previously identified through manual analysis. This opens up opportunities for finding and explaining features in representations that are not derived from SAEs, a direction increasingly supported by recent work \citep{gao2025sparse, arora2026sparse}.

The effectiveness of agentic workflows depends on the underlying LLM: weaker models may fail to meaningfully refine hypotheses, while more capable models produce sharper constraints (Table~\ref{tab:autointerp_refinement_winrates}). This suggests agentic interpretability will improve alongside LLMs, and that high-stakes claims should be grounded in reproducible evidence. However, limitations remain. The refinement loop does not guarantee monotonic convergence, and hypothesis quality depends on the base LLM's ability to generate informative test cases, which is generally weaker for abstract or context-dependent features. Finally, FeatureFinder requires prompt sets with group labels; fully unsupervised discovery remains future work.

\section{Conclusion} \label{sec:conclusion} We present InterpAgent, an agentic framework for automated interpretability that treats feature explanation as an iterative experimental process rather than a one-shot labeling task. Each hypothesis is tied to concrete tests, quantitative outcomes, and an explicit audit trail. By integrating feature discovery, hypothesis refinement, and causal interventions into a single loop accessible through natural language, the system makes interpretability more scalable and usable for both experts and non-experts. As LLMs continue to improve (yielding sharper hypotheses, better test case generation, and more reliable refinement) and as interpretable representations rapidly scale (larger SAEs, weight-sparse models, and emerging feature bases), agentic analysis stands to gain from both directions, potentially enabling systematic auditing of feature libraries at scales that are currently impractical. We see natural extensions to large-scale feature auditing (e.g., re-evaluating repositories such as Neuronpedia), richer causal analyses beyond single-feature interventions, and tighter integration with circuit-level methods that trace not just what a feature represents but how it participates in model computation.

\section*{Code Availability}
The code is available at \url{https://github.com/arnaumarin/InterpAgent}.

\section*{Acknowledgments}
A.ML. acknowledges the support from the RCC-Fellowship of Harvard. 


\bibliography{colm2026_conference}
\bibliographystyle{colm2026_conference}

\appendix
\newpage
\section{Additional Experimental Details}
\label{app:experimental}
\renewcommand{\thetable}{A\arabic{table}}
\setcounter{table}{0}
\renewcommand{\thefigure}{A\arabic{figure}}
\setcounter{figure}{0}

This appendix provides supplementary material for the main text.
Appendix~\ref{subsec:app_metrics} gives formal definitions for the seven evaluation metrics. Appendix~\ref{subsec:app_explainer_details} details the
ranking procedure, stopping conditions, and polysemanticity detection. Appendix~\ref{subsec:app_hyperparameters} reports a hyperparameter sensitivity analysis for FeatureFinder. Appendix~\ref{subsec:app_cost} quantifies computational costs. Appendix~\ref{subsec:app_discovery} provides additional diagnostics for the discovery pipeline (clustering benchmarks and marker retrieval throughput). Appendices~\ref{subsec:app_case1_fulltable} and~\ref{subsec:app_case2_fulltable} report full metric breakdowns for Case Studies 1 and 2, followed by a cross-model causal steering example in Appendix~\ref{subsec:app_spanish}.
Appendix~\ref{subsec:app_mlp_discovery} demonstrates the framework on raw MLP neurons beyond SAE features. Finally,
Appendices~\ref{subsec:app_system_prompt_featurefinder} reproduce the system prompts used in our experiments.
\clearpage

\subsection{Evaluation Metric Definitions}
\label{subsec:app_metrics}

Section~\ref{sec:featureexplainer} introduces seven metrics used to rank hypotheses. Here we provide formal definitions for each. Throughout, let $h$ denote a hypothesis, $\mathcal{P}^{+} = \{x_1^{+}, \ldots, x_n^{+}\}$ the set of activating examples (prompts where the feature fires above a threshold), $\mathcal{P}^{-} = \{x_1^{-}, \ldots, x_n^{-}\}$ the non-activating controls, and $a(x)$ the scalar activation of the target feature on prompt $x$. Each metric returns a value in $[0,1]$ (except $p$-value and Cohen's $d$, which have their natural ranges).

\paragraph{1. Detection F1.}
An LLM classifier receives the hypothesis text and a shuffled mixture of $n$ activating and $n$ non-activating examples (with labels hidden). For each example $x_i$, the LLM predicts \textsc{yes} (matches hypothesis) or \textsc{no}. Let $\hat{y}_i \in \{0,1\}$ be the prediction and $y_i$ the ground-truth label (1 if $x_i \in \mathcal{P}^{+}$). We compute:
\[
\text{Precision} = \frac{TP}{TP + FP}, \quad \text{Recall} = \frac{TP}{TP + FN}, \quad \text{DetF1} = \frac{2 \cdot \text{Precision} \cdot \text{Recall}}{\text{Precision} + \text{Recall}}
\]
where $TP = |\{i : \hat{y}_i = 1 \wedge y_i = 1\}|$, $FP = |\{i : \hat{y}_i = 1 \wedge y_i = 0\}|$, $FN = |\{i : \hat{y}_i = 0 \wedge y_i = 1\}|$. A high DetF1 indicates the hypothesis is discriminative: it accurately predicts which examples activate the feature. We use $n = 5$ examples per class.

\paragraph{2. Fuzzing F1.}
For each activating example, we randomly highlight a subset of tokens and ask the LLM whether the highlighted tokens are the ones responsible for activation according to the hypothesis. Let $s_i \in \{0, 1\}$ be the LLM's judgment for example $i$:
\[
\text{FuzzF1} = \frac{1}{|\mathcal{P}^{+}_{\text{sample}}|}\sum_{i} s_i
\]
This measures robustness: a hypothesis that correctly identifies the activating pattern should survive controlled perturbations. We sample up to 5 examples and highlight 1--2 random tokens per example.

\paragraph{3. Surprisal AUROC.}
The LLM rates how coherent each example is with respect to the hypothesis on a 0--10 scale. Let $c^{+}_i$ and $c^{-}_j$ be the normalized coherence ratings ($\in [0,1]$) for activating and non-activating examples respectively:
\[
\text{SurpAUROC} = \frac{\bar{c}^{+} - \bar{c}^{-} + 1}{2}
\]
where $\bar{c}^{+} = \frac{1}{|\mathcal{P}^{+}_{\text{sample}}|}\sum_i c^{+}_i$ and $\bar{c}^{-} = \frac{1}{|\mathcal{P}^{-}_{\text{sample}}|}\sum_j c^{-}_j$. This discrimination score is bounded in $[0,1]$ with 0.5 indicating no difference.

\paragraph{4. Embedding Similarity.}
Using a sentence embedding model (\texttt{all-MiniLM-L6-v2}), we compute the cosine similarity between the hypothesis text $h$ and each example. Let $\text{sim}(h, x) = \frac{\mathbf{e}_h \cdot \mathbf{e}_x}{\|\mathbf{e}_h\| \|\mathbf{e}_x\|}$ where $\mathbf{e}_h, \mathbf{e}_x$ are the respective embeddings. The discrimination score is:
\[
\text{Embed} = \frac{\overline{\text{sim}}^{+} - \overline{\text{sim}}^{-} + 1}{2}
\]
where $\overline{\text{sim}}^{+} = \frac{1}{n}\sum_{x \in \mathcal{P}^{+}_{\text{sample}}} \text{sim}(h, x)$ and analogously for $\overline{\text{sim}}^{-}$. Values above 0.5 indicate the hypothesis is semantically closer to activating examples than to controls.

\paragraph{5. Statistical Separability ($p$-value).}
We test whether activating examples have significantly higher feature activations than non-activating controls using Welch's $t$-test (unequal variance):
\[
t = \frac{\bar{a}^{+} - \bar{a}^{-}}{\sqrt{s_{+}^2/n_{+} + s_{-}^2/n_{-}}}, \quad p = P(T > t \mid H_0)
\]
where $\bar{a}^{+}, \bar{a}^{-}$ are the mean activations, $s_{+}, s_{-}$ the sample standard deviations, and $n_{+}, n_{-}$ the sample sizes. We use a one-sided alternative ($H_1: \mu_{+} > \mu_{-}$). Lower $p$-values indicate stronger evidence that the hypothesis correctly predicts which prompts produce higher activations.

\paragraph{6. Effect Size (Cohen's $d$).}
The standardized mean difference between activating and non-activating activation distributions:
\[
d = \frac{\bar{a}^{+} - \bar{a}^{-}}{s_{\text{pooled}}}, \quad s_{\text{pooled}} = \sqrt{\frac{(n_{+}-1)s_{+}^2 + (n_{-}-1)s_{-}^2}{n_{+} + n_{-} - 2}}
\]
Cohen's $d$ captures the practical magnitude of separation: $d > 0.2$ is small, $d > 0.5$ is medium, and $d > 0.8$ is large.

\paragraph{7. LLM-as-Judge Coherence.}
An LLM rates how well the hypothesis captures the pattern observed in the top activating examples, on a 0--10 integer scale. The prompt includes the hypothesis text and up to 10 representative activating examples (truncated to 200 characters each). The score is normalized to $[0,1]$:
\[
\text{Judge} = \frac{\text{LLM\_rating}}{10}
\]
This provides a holistic assessment of specificity, consistency with evidence, and overall explanation quality that complements the more targeted metrics above.

\paragraph{Aggregate Ranking.}
Given $M$ metrics and $K$ candidate hypotheses, we compute per-metric ranks $r_m(h_k)$ for each hypothesis $h_k$ (using average rank for ties). The aggregate score is:
\[
\bar{r}(h_k) = \frac{1}{M}\sum_{m=1}^{M} r_m(h_k)
\]
For metrics where lower is better ($p$-value), we rank in ascending order; for all others, we rank in descending order. Hypotheses are then sorted by $\bar{r}$, with ties broken by Pareto dominance: $h_1$ dominates $h_2$ if $r_m(h_1) \leq r_m(h_2)$ for all $m$ and $r_m(h_1) < r_m(h_2)$ for at least one $m$. In our experiments, we use $n = 24$ test cases per evaluation (distributed across positive, negative, edge, and adversarial categories) and $M = 7$ metrics.

\clearpage
\subsection{FeatureExplainer: additional details}
\label{subsec:app_explainer_details}

This section provides full details on the components summarized in
Section~\ref{sec:featureexplainer}.

\paragraph{Multi-objective ranking.}
Hypotheses are assessed along the seven-metric battery (defined in
Appendix~\ref{subsec:app_metrics}); for each metric $m$, candidates are ranked
to obtain a per-metric ordinal score $r_m(h)$, with ties handled by averaging.
An aggregate score is computed by averaging ordinal ranks across metrics. In
parallel, Pareto dominance filters candidates: $h_1$ dominates $h_2$ if it is
at least as good on every metric and strictly better on at least one. We
prioritize hypotheses on the Pareto frontier by increasing aggregate rank. The
ranking is used diagnostically: poor Detection suggests an overly broad
trigger; poor Fuzzing indicates spurious cues; weak separability suggests
unreliable activation prediction; low semantic alignment indicates
underspecification. Revised hypotheses are constrained to be diverse from prior
attempts, both semantically and structurally.

\paragraph{Stopping conditions.}
We declare convergence when (i) the top-ranked hypothesis remains unchanged
for consecutive iterations, (ii) the aggregate rank improves by less than
$\epsilon = 0.5$ rank units, and (iii) newly generated negative controls do
not reveal additional counterexamples (no increase in false positives under
Detection/Fuzzing). At termination, the agent returns the best-supported
hypothesis with the full metric profile and an audit trail.

\paragraph{Polysemanticity detection.}
We collect high-performing hypotheses satisfying both an aggregate-rank
threshold and a minimum coherence score. Semantic distinctness is measured
between hypotheses using a judge model (normalized to $[0,1]$). A feature is
flagged as polysemantic when at least two semantically distinct hypotheses
exceed the support thresholds and neither dominates the other across metrics.
In this case, all competing interpretations and their evidence profiles are
reported.

\paragraph{Per-iteration refinement breakdown.}
Table~\ref{tab:autointerp_refinement_winrates_full} provides the full
per-layer and per-iteration breakdown summarized in
Table~\ref{tab:autointerp_refinement_winrates} of the main text. Each row
sums to 100\%: the Baseline column shows how often the one-shot label remains
top-ranked, and Refined-$k$ columns show how often the hypothesis produced at
iteration $k$ is selected as best. The gap between stronger models (Gemini 2.5
Pro, Claude 4 Sonnet) and weaker models (GPT-4o mini) is consistent across
both layers.

\begin{table}[t]
\centering
\renewcommand{\arraystretch}{1.15}
\setlength{\tabcolsep}{5pt}
\caption{\textbf{Per-iteration refinement win rates (\% of features).}
Baseline is a one-shot autointerp label from
\texttt{pile-uncopyrighted}; Refined-$k$ is the best hypothesis after $k$
refinement iterations. Each row sums to 100\%.}
\label{tab:autointerp_refinement_winrates_full}
\resizebox{\columnwidth}{!}{%
\begin{tabular}{l c c c c c r}
\toprule
Run & Baseline & Refined-1 & Refined-2 & Refined-3 & Refined-4 & $N$ features \\
\midrule
\multicolumn{7}{l}{\textbf{Layer 17}} \\
\midrule
Claude 4 Sonnet & 16.7 & 16.7 & 21.7 & 21.7 & 23.3 & 60 \\
Gemini 2.5 Pro  & 20.5 & 24.5 & 17.5 & 17.5 & 20.0 & 200 \\
GPT-4o mini     & 34.0 &  6.0 & 18.0 & 19.0 & 23.0 & 100 \\
\midrule
\multicolumn{7}{l}{\textbf{Layer 13}} \\
\midrule
Gemini 2.5 Pro  & 21.0 & 27.0 & 18.0 & 21.0 & 13.0 & 100 \\
GPT-4o mini     & 39.1 &  9.1 & 17.6 & 16.2 & 18.1 & 1{,}500 \\
\bottomrule
\end{tabular}%
}
\end{table}

\clearpage
\subsection{Hyperparameter sensitivity analysis}
\label{subsec:app_hyperparameters}

FeatureFinder's discovery pipeline depends on three key hyperparameters:
(1)~the number of PCA components used to construct the activation embedding,
(2)~the number of neighbors $k$ in the $k$-NN graph, and (3)~the Leiden
clustering resolution. Here we report the default values, provide a systematic
ablation, and assess the sensitivity of downstream discovery quality to each
parameter.

\paragraph{Default configuration.}
The pipeline proceeds as follows (see Section~\ref{sec:featurefinder}):
\begin{enumerate}
\item Per-prompt SAE activation vectors are log-transformed and
      per-prompt normalized.
\item PCA is applied with up to $\min(200,\, N{-}1)$ components,
      where $N$ is the number of prompts.
\item A $k$-nearest-neighbor graph is constructed using
      $\min(n_{\text{pcs}}{-}5,\, 50)$ principal components, with $k = 15$.
\item Leiden community detection is applied at resolution $0.5$.
\item Wilcoxon rank-sum tests with Benjamini--Hochberg correction
      identify marker features per cluster. Candidates are retained if
      adjusted $p < 0.001$, score $> 2.0$, log-fold change $> 0.5$,
      and Cohen's $d > 0.3$.
\end{enumerate}

\paragraph{Ablation setup.}
We sweep over PCA dimensions $\in \{50, 200\}$, $k \in \{10, 15, 25\}$, and
Leiden resolution $\in \{0.3, 0.5, 1.0\}$, yielding 18 configurations. Each
configuration is evaluated on 17 languages with 200 prompts per language at
layer~0, measuring clustering agreement against ground-truth language labels
using Adjusted Rand Index (ARI), Normalized Mutual Information (NMI),
V-measure, and Silhouette score. Table~\ref{tab:app_ablation_full} reports all
configurations sorted by ARI, with the paper default marked.

\begin{table}[h]
\centering
\small
\caption{\textbf{Full hyperparameter ablation.} Clustering quality across 18
configurations (17 languages, 200 prompts each, layer~0). $^\dagger$Paper
default.}
\label{tab:app_ablation_full}
\begin{tabular}{rrrccccr}
\toprule
PCA & $k$ & Res. & ARI & NMI & V-meas & Silh. & Clusters \\
\midrule
 50 & 25 & 0.3 & \textbf{0.031} & 0.155 & 0.155 & 0.390 & 185 \\
 50 & 15 & 0.3 & 0.030 & 0.161 & 0.161 & 0.340 & 248 \\
200 & 15 & 0.3 & 0.030 & 0.162 & 0.162 & 0.326 & 244 \\
200 & 25 & 0.3 & 0.030 & 0.151 & 0.151 & 0.397 & 187 \\
200 & 25 & 0.5 & 0.029 & 0.162 & 0.162 & \textbf{0.428} & 204 \\
 50 & 25 & 0.5 & 0.026 & 0.161 & 0.161 & 0.404 & 206 \\
\midrule
200 & 15 & 0.5 & 0.024 & \textbf{0.167} & \textbf{0.167} & 0.350 & 271$^\dagger$ \\
\midrule
 50 & 10 & 0.3 & 0.023 & 0.164 & 0.164 & 0.293 & 321 \\
200 & 10 & 0.3 & 0.023 & 0.165 & 0.165 & 0.315 & 315 \\
200 & 25 & 1.0 & 0.023 & 0.167 & 0.167 & 0.341 & 243 \\
 50 & 15 & 0.5 & 0.022 & 0.167 & 0.167 & 0.348 & 274 \\
 50 & 25 & 1.0 & 0.022 & 0.168 & 0.168 & 0.354 & 249 \\
 50 & 10 & 0.5 & 0.021 & 0.168 & 0.168 & 0.326 & 341 \\
200 & 10 & 0.5 & 0.020 & 0.169 & 0.169 & 0.319 & 345 \\
200 & 15 & 1.0 & 0.019 & 0.174 & 0.174 & 0.339 & 317 \\
 50 & 15 & 1.0 & 0.018 & 0.171 & 0.171 & 0.317 & 317 \\
 50 & 10 & 1.0 & 0.017 & 0.174 & 0.174 & 0.306 & 396 \\
200 & 10 & 1.0 & 0.016 & 0.175 & 0.175 & 0.331 & 396 \\
\bottomrule
\end{tabular}
\end{table}

\begin{figure}[h]
\centering
\includegraphics[width=\textwidth]{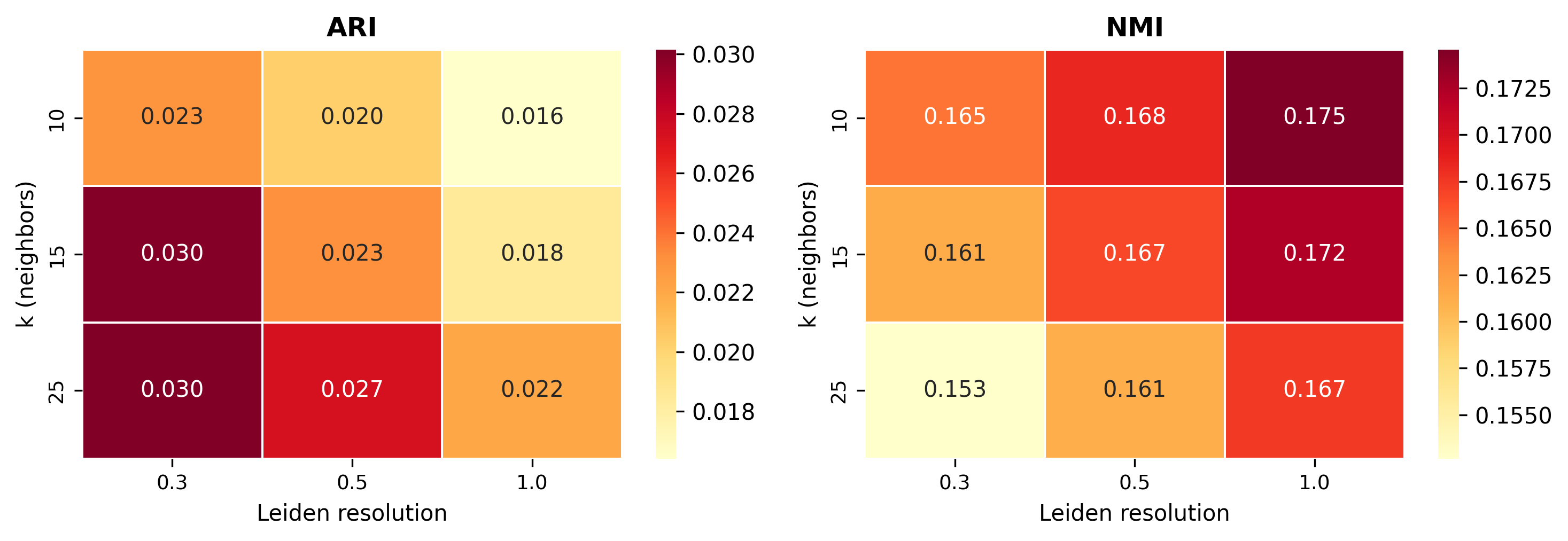}
\caption{\textbf{Hyperparameter sensitivity: clustering quality.}
ARI and NMI heatmaps across $k$ and Leiden resolution (averaged over PCA
dimensions). NMI varies by less than 13\% across all 18 configurations.}
\label{fig:app_ablation_heatmap}
\end{figure}

\paragraph{Trends.}
Three patterns emerge from the ablation
(Figure~\ref{fig:app_ablation_heatmap}). First, \emph{resolution controls the
ARI/NMI trade-off}: lower resolution ($0.3$) produces fewer, coarser clusters
that better match the 17 ground-truth labels, yielding higher ARI, while
higher resolution ($1.0$) produces finer-grained clusters with higher NMI as
sub-language distinctions emerge. The default of $0.5$ balances these
objectives. Second, \emph{larger $k$ favors ARI}: increasing $k$ from 10 to 25
improves ARI ($0.020 \to 0.027$ on average) by producing smoother graphs that
merge nearby clusters, at a slight cost to NMI ($0.169 \to 0.160$). Third,
\emph{PCA dimensionality has minimal effect}: changing from 50 to 200
components shifts ARI by less than 3\% and NMI by less than 0.1\%, indicating
that the top 50 principal components already capture the relevant variance.

Overall, clustering quality is robust to hyperparameter choice: NMI varies by
less than 13\% ($0.151$--$0.175$) across all 18 configurations. The paper
defaults ($k = 15$, resolution $= 0.5$, PCA $\leq 200$) achieve the highest
NMI among same-resolution configurations and lie within 22\% of the best ARI.
We note that absolute ARI values in this ablation (${\sim}0.03$) are lower than
Table~\ref{tab:app_clustering_metrics} ($0.315$) due to reduced data scale
(200 vs.\ 500 prompts per language, 17 vs.\ 20 languages); the ablation
demonstrates relative robustness, not absolute performance.

\begin{figure}[h]
\centering
\includegraphics[width=0.75\textwidth]{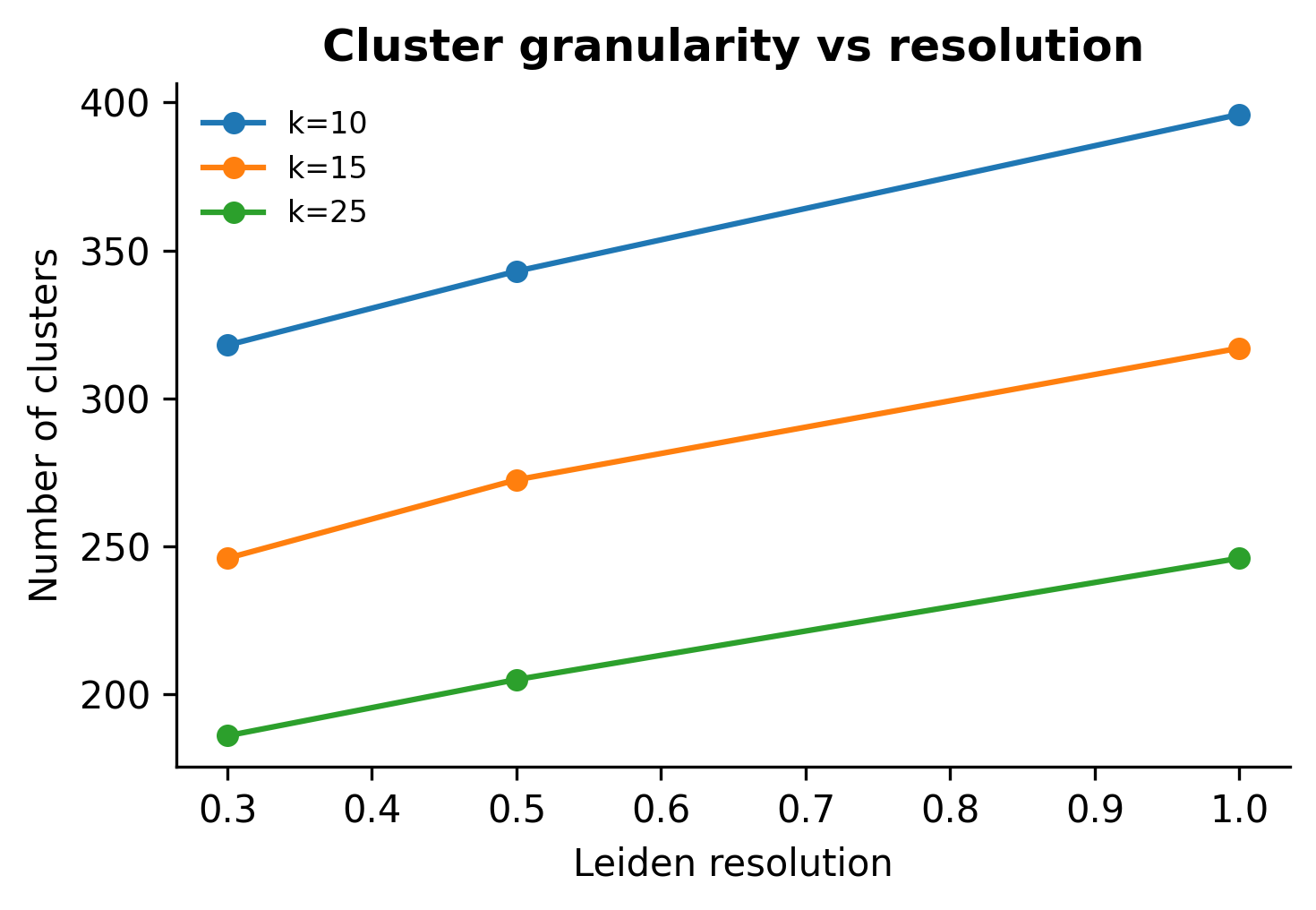}
\caption{\textbf{Hyperparameter sensitivity: cluster count.}
Number of clusters vs.\ Leiden resolution. Higher resolution produces
finer-grained clusters; larger $k$ reduces total cluster count.}
\label{fig:app_ablation_clusters}
\end{figure}

\clearpage

%

\subsection{Cost analysis}
\label{subsec:app_cost}

We quantify the computational overhead of InterpAgent's iterative refinement loop relative to a standard one-shot autointerp baseline. Both methods were run on the same 26 SAE features using \texttt{gpt-4o-mini} as the hypothesis generator. The baseline generates a single hypothesis from top-activating examples in one LLM call; InterpAgent runs four refinement iterations,
generating test cases, collecting SAE activations, obtaining LLM criticism, and revising hypotheses at each step.

Table~\ref{tab:cost} reports the per-feature cost breakdown. InterpAgent requires 13 LLM calls and approximately 17{,}000 tokens per feature, resulting in a 34$\times$ cost overhead. In absolute terms, however, the cost remains negligible: explaining a single feature costs \$0.006, and a full sweep of 1{,}000 features costs approximately \$6. The additional expense buys iterative hypothesis testing with real SAE activations, an audit trail of tested alternatives, and polysemanticity detection---none of which are available from one-shot labeling.

To assess how these costs scale with model capability, Table~\ref{tab:cost_models} projects InterpAgent's cost across five model tiers based on published API pricing. Even with frontier models, the cost per feature remains under \$0.13,
and a 1{,}000-feature sweep stays below \$125. For most practical
applications, mid-tier models such as \texttt{gpt-4o-mini} offer the best trade-off between explanation quality and cost.

\begin{table}[t]
\centering
\caption{\textbf{Cost comparison: baseline one-shot autointerp vs. InterpAgent iterative refinement.} Measured with \texttt{gpt-4o-mini}. Token counts from instrumented API tracking; wall-clock times and costs averaged over
26~features.}
\label{tab:cost}
\begin{tabular}{lrrrr}
\toprule
Method & Calls & Tokens & Time (s) & Cost (\$) \\
\midrule
Baseline (1-shot)       &  1 &      727 &   7.4 & 0.0002 \\
InterpAgent (iterative) & 13 & 16{,}867 & 176.3 & 0.0061 \\
\midrule
Overhead & 13$\times$ & 23$\times$ & 24$\times$ & 34$\times$ \\
\bottomrule
\end{tabular}
\end{table}

\begin{table}[t]
\centering
\caption{\textbf{Estimated InterpAgent cost per feature across model tiers}
(13~LLM calls, ${\sim}$765 input/output tokens each). Even with frontier
models, the cost per feature remains under \$0.13.}
\label{tab:cost_models}
\begin{tabular}{lrrrr}
\toprule
Model & Input/1M\$ & Output/1M\$ & Per feat.\ (\$) & 1{,}000 feat.\ (\$) \\
\midrule
\texttt{gpt-4o-mini}  &  0.15 &   0.60 & 0.002 &    1.86 \\
\texttt{gpt-5.4-nano} &  0.20 &   0.80 & 0.003 &    2.49 \\
\texttt{gpt-5.4-mini} &  1.00 &   4.00 & 0.012 &   12.43 \\
\texttt{gpt-4o}       &  2.50 &  10.00 & 0.031 &   31.07 \\
\texttt{gpt-5.4}      & 10.00 &  40.00 & 0.124 &  124.28 \\
\bottomrule
\end{tabular}
\end{table}

\begin{table*}[t]
\centering
\caption{\textbf{Clustering performance on SAE activation space} (agreement
vs.\ reference language labels, 20~languages, 500~prompts each, layer~0). The
$k$-NN graph with Leiden community detection outperforms global partitioning
baselines across all metrics.}
\label{tab:app_clustering_metrics}
\begin{tabular}{lcccccc}
\toprule
Method & ARI & NMI & Homogeneity & Completeness & V-measure & Silhouette \\
\midrule
$k$-NN graph   & 0.315 & 0.536 & 0.548 & 0.525 & 0.536 &  0.005 \\
Agglomerative  & 0.130 & 0.304 & 0.282 & 0.329 & 0.304 &  0.078 \\
K-Means        & 0.118 & 0.278 & 0.264 & 0.293 & 0.278 &  0.112 \\
Random         & 0.000 & 0.007 & 0.007 & 0.007 & 0.007 & $-$0.042 \\
\bottomrule
\end{tabular}
\end{table*}


\begin{figure*}[t]
\centering
\includegraphics[width=\linewidth]{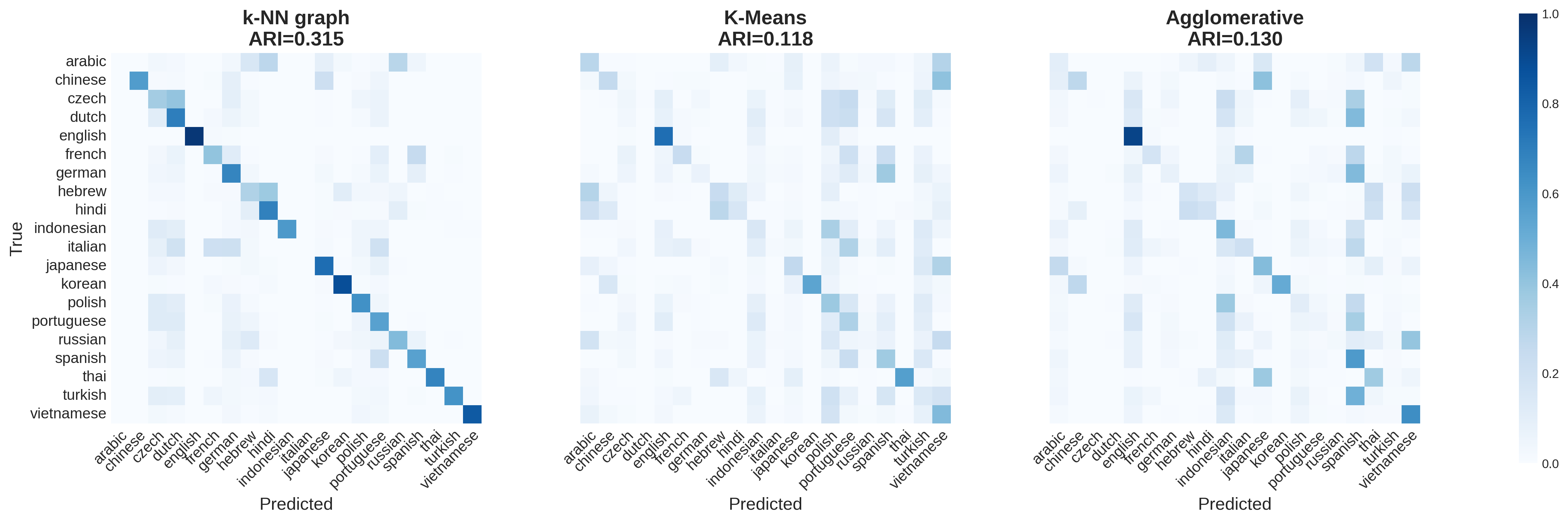}
\caption{\textbf{Normalized confusion matrices for the clustering benchmark}
(20~languages, 500~prompts each, layer~0). Clusters are aligned to the closest
matching language labels before plotting. \textbf{Left:} $k$-NN graph with
Leiden community detection (ARI = 0.315). \textbf{Center:} K-Means
(ARI = 0.118). \textbf{Right:} Agglomerative with Ward linkage
(ARI = 0.130).}
\label{fig:app_clustering_comparison}
\end{figure*}

\newpage

\subsection{Feature discovery: additional diagnostics}
\label{subsec:app_discovery}

We report supporting experiments for FeatureFinder that validate the clustering prior and characterize marker retrieval throughput.

\paragraph{Clustering benchmark.}
We first ask whether neighborhood structure in SAE activation space is sufficient to recover language labels in a controlled setting. We generate 500~prompts per language for 20~languages (10{,}000 prompts total): Arabic, Chinese, Czech, Dutch, English, French, German, Hebrew, Hindi, Indonesian, Italian, Japanese, Korean, Polish, Portuguese, Russian, Spanish, Thai, Turkish,
and Vietnamese. Each prompt is represented by its SAE activation vector at layer~0. We compare our $k$-NN graph with Leiden community detection against K-Means, Agglomerative clustering, and random assignment as a lower bound. Because cluster identities are arbitrary, we align clusters to the closest
matching language labels before evaluation. Table~\ref{tab:app_clustering_metrics} reports agreement metrics and Figure~\ref{fig:app_clustering_comparison} shows the corresponding normalized confusion matrices. The $k$-NN graph substantially outperforms both baselines across all agreement metrics (ARI~0.315 vs.\ 0.130 and 0.118), confirming that
local neighborhood structure in activation space is better suited for category-level discovery than global partitioning methods.

\paragraph{Marker retrieval throughput.}
We evaluate how many candidate markers the discovery pipeline produces and how many survive automated screening. For the language benchmark, we use 18~languages grouped into five families: Romance (French, Spanish, Italian, Portuguese), Germanic (English, German, Dutch), Slavic (Russian, Polish,
Czech), Semitic (Arabic, Hebrew), and East Asian (Chinese, Japanese, Korean). Table~\ref{tab:app_family_candidates_judge} reports the resulting counts. Only a small fraction of candidates are retained (0.2--4.0\%), reflecting the deliberate design trade-off between permissive initial retrieval and aggressive screening. For the non-linguistic benchmark, we run FeatureFinder across layers~0, 8, and 11 on three categories: coding, math, and safety. Table~\ref{tab:nonling_markers} reports the per-layer breakdown. The layer-specific distribution of markers is consistent with the expectation that surface-level token cues are captured early while syntactic patterns emerge at intermediate depths.


\begin{table}[t]
\centering
\caption{\textbf{Language-family marker retrieval summary.} ``Found'' counts all candidate markers returned by the statistical discovery pipeline. ``Validated'' counts the subset retained after LLM-as-judge screening for specificity and relevance. For the language-family marker discovery was performed in SAE layer 0.} \label{tab:app_family_candidates_judge}
\begin{tabular}{lrrr}
\toprule
Family & Found & Validated & \% \\
\midrule
Germanic   &    753 & 30 & 4.0 \\
Romance    & 2{,}453 & 18 & 0.7 \\
Slavic     &    571 &  8 & 1.4 \\
Semitic    & 3{,}394 &  6 & 0.2 \\
East Asian & 1{,}182 &  7 & 0.6 \\
\bottomrule
\end{tabular}
\end{table}


\begin{table}[t]
\centering
\caption{\textbf{Non-linguistic marker features discovered per category and layer.} FeatureFinder was run on \texttt{gemma-2-2b} SAE features at layers~0, 8, and 11 with strict statistical thresholds (adjusted $p < 0.001$, Cohen's $d > 0.3$, log-fold change $> 0.5$). Dashes indicate zero markers validated.}
\label{tab:nonling_markers}
\begin{tabular}{lccc}
\toprule
Category & Layer 0 & Layer 8 & Layer 11 \\
\midrule
Math (symbolic notation) & 3 & 1 & --- \\
Coding (Python prompts)  &  3 & 2 & --- \\
Safety (harmful)  &  --- & 1 & 3 \\
\bottomrule
\end{tabular}
\end{table}

\clearpage

\subsection{Case Study 1: Full hypothesis metrics (French feature)}
\label{subsec:app_case1_fulltable}

The main text summarizes the top hypotheses for Feature {L0\_F9004}. Table~\ref{tab:app_L0_9004_full} reports the full metric breakdown used by the agent for ranking. ``Avg.\ Rank'' is the mean ordinal rank across the scored metrics (lower is better).

\begin{table*}[t]
    \centering
    \scriptsize
    \renewcommand{\arraystretch}{1.12}
    \caption{\textbf{Case Study 1 (Feature {L0\_F9004})}. Full hypothesis metrics and aggregate ranking.}
    \label{tab:app_L0_9004_full}
    \resizebox{\textwidth}{!}{%
    \begin{tabular}{r l p{6.8cm} c c c c c c c c}
        \toprule
        \textbf{Rank} & \textbf{Source} & \textbf{Hypothesis} &
        \textbf{Avg.\ Rank} &
        \textbf{Det.\ F1} & \textbf{Fuzz F1} & \textbf{Surp.\ AUROC} &
        \textbf{Embed} & \textbf{Judge} &
        \textbf{$p$-val} & \textbf{$d$} \\
        \midrule
        1 & Refined-2 &
        The feature identifies informal French syntax and phrases, particularly in contexts discussing technology, troubleshooting, and online communication, emphasizing colloquial and pragmatic language use. &
        1.92 &
        1.00 & 0.75 & 0.71 &
        0.59 & 0.40 &
        {$<0.001$} & 2.53 \\
        2 & Refined-1 &
        The feature detects French colloquial expressions and syntax patterns commonly used in informal discussions about technology and troubleshooting. &
        2.28 &
        1.00 & 0.00 & 0.75 &
        0.58 & 0.60 &
        {$<0.001$} & 2.24 \\
        3 & Refined-4 &
        The feature detects informal French syntax and colloquial expressions focusing on technology-related discussions and pragmatic usage. &
        3.21 &
        1.00 & 0.00 & 0.66 &
        0.58 & 0.40 &
        0.026 & 1.39 \\
        4 & Refined-3 &
        The feature detects informal French syntax and colloquial phrases, focusing on technology-related discussions and troubleshooting. &
        3.50 &
        1.00 & 0.00 & 0.63 &
        0.59 & 0.40 &
        0.879 & -0.72 \\
        5 & Neuronpedia &
        Expressions related to knowledge and understanding, particularly about technology and troubleshooting. &
        4.07 &
        0.89 & 0.00 & 0.77 &
        0.52 & 0.00 &
        1.00 & 0.00 \\
        \bottomrule
    \end{tabular}%
    }
\end{table*}

\begin{figure}[t]
    \centering
    \includegraphics[width=0.65\linewidth]{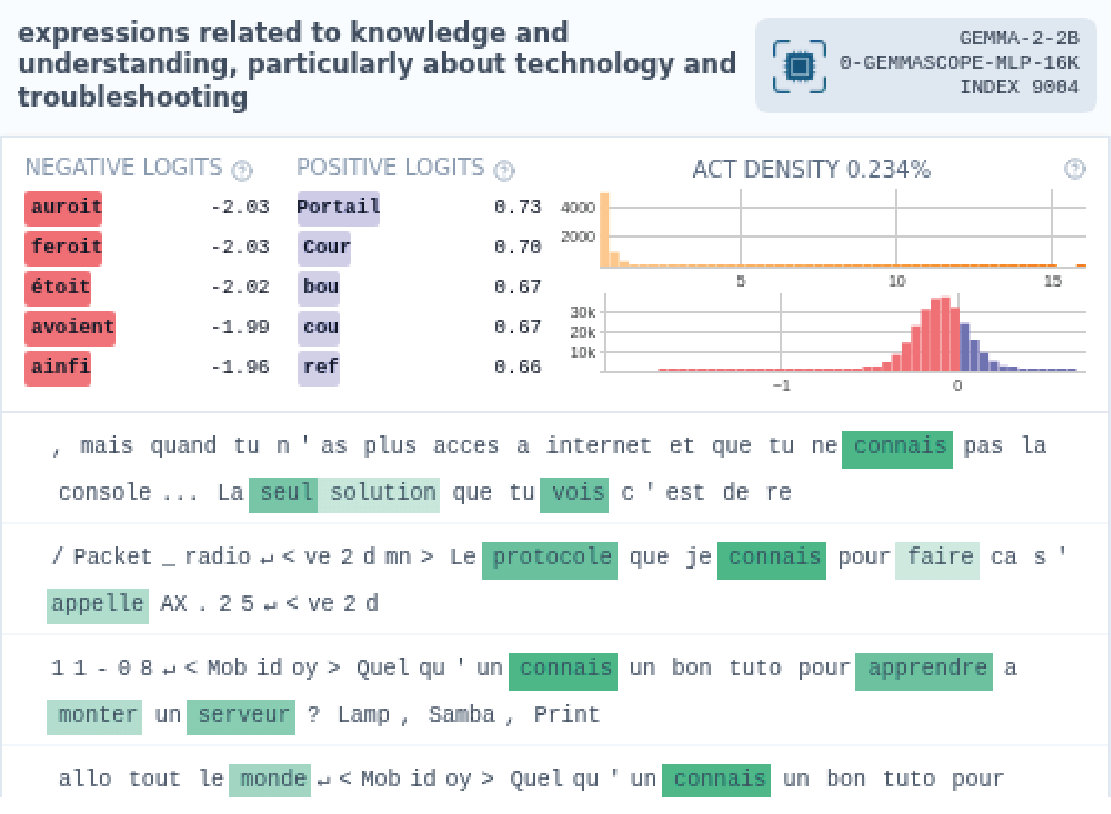}
    \caption{\textbf{Feature L0\_F9004 in Neuronpedia.} Representative top-activating examples (highlighted tokens) alongside the baseline one-shot auto-interpretation produced with GPT-4o-mini.}
    \label{fig:neuronpedia_L0_F900_UX}
\end{figure}

\begin{figure}[t]
    \centering
    \includegraphics[width=\linewidth]{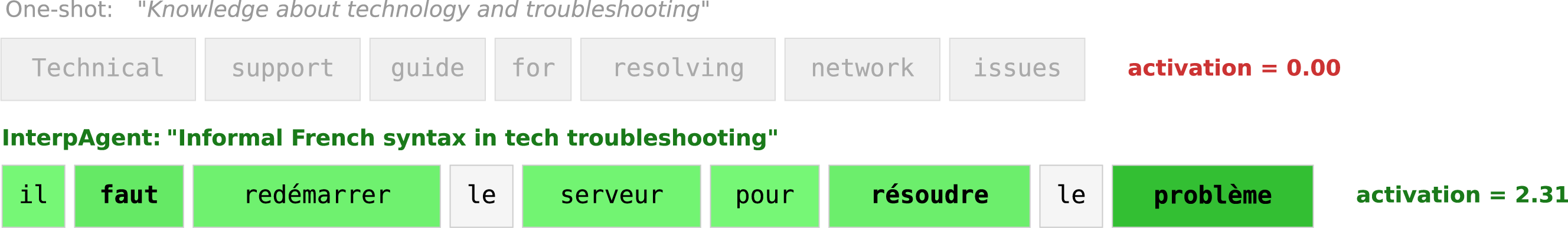}
    \caption{\textbf{Feature L0\_F9004.} Representative high-activating examples (highlighted tokens) alongside the baseline one-shot auto-interpretation produced with GPT-4o-mini.}
    \label{fig:neuronpedia_L0_F9004}
\end{figure}

\clearpage

\subsection{Case Study 2: Full hypothesis metrics (polysemantic feature)}
\label{subsec:app_case2_fulltable}

The main text reports a compact ranking for Feature {L4\_F12773}. Table~\ref{tab:app_L4_12773_full} contains the full metrics used in the ranking. As above, ``Avg.\ Rank'' is the mean ordinal rank across the scored metrics.

\begin{table*}[t]
    \centering
    \scriptsize
    \renewcommand{\arraystretch}{1.12}
    \caption{\textbf{Case Study 2 (Feature {L4\_F12773})}. Full hypothesis metrics and aggregate ranking.}
    \label{tab:app_L4_12773_full}
    \resizebox{\textwidth}{!}{%
    \begin{tabular}{r l p{6.8cm} c c c c c c c c}
        \toprule
        \textbf{Rank} & \textbf{Source} & \textbf{Hypothesis} &
        \textbf{Avg.\ Rank} &
        \textbf{Det.\ F1} & \textbf{Fuzz F1} & \textbf{Surp.\ AUROC} &
        \textbf{Embed} & \textbf{Judge} &
        \textbf{$p$-val} & \textbf{$d$} \\
        \midrule
        1 & FeatureFinder &
        Coding snippets and equations &
        2.286 &
        1.00 & 1.00 & 0.79 &
        1.00 & 0.60 &
        {$<0.001$} & 13.07 \\
        2 & Autointerp &
        Terms related to medical research and methodologies &
        2.571 &
        1.00 & 1.00 & 0.96 &
        1.00 & 0.80 &
        0.064 & 0.99 \\
        3 & Refined-1 &
        Terms related to pharmacology and geometry, with a focus on compound words or hyphenated forms in scientific discourse. &
        3.357 &
        1.00 & 1.00 & 0.53 &
        0.88 & 0.70 &
        {$<0.001$} & 8.54 \\
        4 & Refined-2 &
        Terms related to pharmacology and geometry, focusing on English scientific compound words and occasionally displaying Spanish morphological patterns in complex noun phrases. &
        3.714 &
        1.00 & 1.00 & 0.57 &
        0.90 & 0.40 &
        0.038 & 1.22 \\
        5 & Refined-4 &
        Scientific discourse featuring pharmacological and geometric terms, highlighting English compound forms alongside Spanish morphology in compound nouns and noun-adjective agreements. &
        3.786 &
        1.00 & 1.00 & 0.58 &
        0.95 & 0.80 &
        0.182 & 0.55 \\
        6 & Refined-3 &
        Scientific discourse featuring pharmacological and geometric terms, emphasizing English compound forms and occasional Spanish morphology, with specific focus on noun-adjective agreement patterns. &
        5.286 &
        0.80 & 0.80 & 0.60 &
        0.83 & 0.30 &
        0.088 & 0.86 \\
        \bottomrule
    \end{tabular}%
    }
\end{table*}

\clearpage


\subsection{Additional case study: Cross-model causal persistence (Spanish feature)}
\label{subsec:app_spanish}

This example follows a refined hypothesis with a causal check. Feature L0\_{11826} is interpreted as Spanish or Hispanic language and cultural cues. We apply the same activation intervention to the base model (\texttt{gemma-2-2b}) and its instruction-tuned variant (\texttt{gemma-2-2b-it}). Figure~\ref{fig:app_spanish_steering} shows a consistent, hypothesis-aligned shift in both settings, suggesting the direction remains usable after instruction tuning.

\begin{figure*}[t]
    \centering
    \small
    \renewcommand{\arraystretch}{1.25}
    \caption{\textbf{Cross-model persistence of Feature L0\_{11826}}. The same activation intervention induces Spanish code-switching and culturally specific content in both the base and instruction-tuned models.}
    \label{fig:app_spanish_steering}

    \textbf{(A) Base model (\texttt{gemma-2-2b})} \\[2pt]
    \begin{tabular}{p{0.25\linewidth} p{0.32\linewidth} p{0.35\linewidth}}
        \toprule
        \textbf{Input} & \textbf{Baseline} & \textbf{Steered (+50)} \\
        \midrule
        \textit{"Hello, my name is"} &
        Hello, my name is \textbf{Michael} and I am a student at the \textbf{University of Iowa}... &
        Hello, my name is \textcolor{steer_red}{\textbf{Carlos}} and I am going to study... \textcolor{steer_red}{\textbf{University of Costa Rica}}... \\
        \bottomrule
    \end{tabular}

    \vspace{0.25cm}

    \textbf{(B) Instruction-tuned model (\texttt{gemma-2-2b-it})} \\[2pt]
    \begin{tabular}{p{0.25\linewidth} p{0.32\linewidth} p{0.35\linewidth}}
        \toprule
        \textbf{Input} & \textbf{Baseline} & \textbf{Steered (+50)} \\
        \midrule
        \textit{"What food should I cook today?"} &
        I suggest a delicious \textbf{chicken stir-fry}! &
        Here are some ideas: \textbf{\textcolor{steer_red}{Pasta:}} \textcolor{steer_red}{una receta...} \ \textbf{\textcolor{steer_red}{tacos:}} \textcolor{steer_red}{una...} \\
        \midrule
        \textit{"What country would you recommend for a holiday?"} &
        I would recommend \textbf{Italy}. &
        I recommend \textcolor{steer_red}{\textbf{España}}. \\
        \bottomrule
    \end{tabular}
\end{figure*}

\clearpage

\subsection{Beyond SAE features: MLP neuron discovery in weight-sparse transformers}
\label{subsec:app_mlp_discovery}

While our main experiments use SAE features as the analysis basis, the InterpAgent framework is not restricted to any particular feature representation. To demonstrate this generality, we apply FeatureFinder's statistical discovery pipeline directly to raw MLP neuron activations in a weight-sparse transformer from \citet{gao2025sparse}, without any SAE.

\paragraph{Setup.} We use their smaller model, a 12-layer GPT-2-style transformer with 1{,}024-dimensional hidden states and 4{,}096 MLP neurons per layer, trained on a Python code dataset with aggressive weight sparsity ($L_0$ constraint). We extract post-GELU MLP activations directly and treat each of the 4{,}096 neurons as a candidate feature. We construct 16 prompt categories covering Python syntax constructs (e.g., single/double-quote strings, import statements, function calls, class definitions, try/except blocks, comments, decorators) with 100 prompts per category (1{,}600 total). FeatureFinder's statistical pipeline is applied identically to the SAE experiments, substituting raw neuron activations for SAE feature activations.

\paragraph{Discovery results.} The pipeline discovers statistically significant marker neurons for all 16 categories, with effect sizes often an order of magnitude greater than those observed for SAE features, reflecting the monosemantic nature of neurons in weight-sparse models. Table~\ref{tab:app_mlp_selected} highlights a representative subset including the two neurons that match ground truth from \citet{gao2025sparse}.

\begin{table}[h]
\centering
\small
\caption{\textbf{Selected MLP marker neurons discovered (layer~0).} Representative subset of 16 discovered categories. $^\dagger$Neurons identified by \citet{gao2025sparse} via manual circuit analysis.}
\label{tab:app_mlp_selected}
\begin{tabular}{lrrcc}
\toprule
Category & Neuron & $d$ & Accuracy & AUROC \\
\midrule
Double-quote strings$^\dagger$ &  863 &  73.9 & --- & --- \\
Single-quote strings$^\dagger$ & 2790 &  13.7 & --- & --- \\
\midrule
Try/except           &  615 &  17.0 & 1.000 & --- \\
Import statements    & 2043 &  36.4 & 0.938 & 1.000 \\
Function calls       & 4020 &  13.4 & 0.938 & 0.867 \\
Semicolon ending     & 3342 & 152.3 & 0.875 & --- \\
Comments (\texttt{\#}) & 2082 & 72.3 & 0.750 & 1.000 \\
Class definitions    &  831 &  60.6 & 0.688 & 0.836 \\
\midrule
\textbf{Mean (14 validated through FeatureExplainer)} & & & \textbf{0.750} & \textbf{0.893} \\
$>$50\% accuracy     & & & \multicolumn{2}{c}{12/14 (86\%)} \\
\bottomrule
\end{tabular}
\end{table}

\paragraph{Agreement with manual circuit analysis.} \citet{gao2025sparse} manually analyzed circuits underlying simple Python tasks in the sparse model, identifying four key neurons in layer~0 responsible for string-closing behavior: neuron~985 (fires on both quote types), neuron~460 (positive on double quotes, negative on single), neuron~863 (combined detector/classifier), and neuron~2790 (single-quote detector). Our automated pipeline rediscovers two of these four without any task-specific supervision: neuron~863 as the top marker for double-quote strings ($d = 73.9$), and neuron~2790 as a top-ranked marker for single-quote strings ($d = 13.7$). The two neurons not recovered are expected omissions: neuron~985 fires on both quote types and therefore does not discriminate between our categories, while neuron~460 operates on the residual stream delta rather than the post-GELU activations we measure. Notably, this provides validation against human expert judgments: neurons 863 and 2790, which our pipeline identifies as top markers for their respective categories, were independently discovered by the paper through manual circuit analysis requiring approximately one researcher-day, suggesting that the agent's discovery aligns with human expert findings, significantly faster (see \ref{subsec:app_cost} for references).

\paragraph{Validation with hypothesis testing.} To confirm that the discovered markers are not statistical artifacts, we run FeatureExplainer's hypothesis-test-refine loop on the top neuron from 14 categories. For each neuron, an LLM generates a natural-language hypothesis, then produces positive and negative test prompts evaluated against real MLP activations. The mean validation accuracy is 75\%, with 12 of 14 neurons (86\%) exceeding chance level (Table~\ref{tab:app_mlp_selected}). Categories with unambiguous syntactic markers (try/except, import, function calls) achieve the highest accuracy, while more context-dependent categories (return statements, numeric operations) prove harder for LLM-generated test cases to capture.

\paragraph{Implications.} These results demonstrate that InterpAgent's discovery and explanation pipeline applies to any neural representation where per-prompt activations can be measured. The same statistical machinery transfers directly to raw MLP neurons, achieving high validation accuracy (75\%) because activation boundaries in weight-sparse models are cleaner. The pipeline automatically recovers neuron-level findings that required manual circuit analysis in prior work, reducing analysis time from approximately one researcher-day to under two minutes.
\clearpage


\FloatBarrier
\subsection{System prompt used for InterpAgent}
\label{subsec:app_system_prompt_featurefinder}

For reproducibility, we include the (verbatim) system prompt used to run the InterpAgent agent in our experiments.

\thispagestyle{plain}

\begin{tcolorbox}[
  enhanced,
  breakable,
  colback=white,
  colframe=black!60,
  boxrule=0.6pt,
  arc=1.5mm,
  left=3mm,right=3mm,top=2mm,bottom=2mm,
  title=\textbf{System prompt for InterpAgent.} Verbatim prompt used to run the refinement and hypothesis-testing loop.
]
\footnotesize
\begin{verbatim}
prompt_template = """
You are an Interp agent responsible for orchestrating mechanistic interpretability 
workflows for language models.

You coordinate specialized subagents and provide them with necessary commands and 
responses.
Each subagent has its own pipeline and will ask you for specific information when 
needed.

**When you response to your subagent, always call the subagent tool with your 
response!!**

=============================
GENERAL INSTRUCTIONS
=============================
- Observe the pipeline and ensure subagents execute their respective steps
- NEVER create or invent new tasks on your own
- NEVER reassign tasks creatively - each subagent is responsible for its own domain
- If user wants end-to-end process, run entire pipeline without asking for confirmations
---

## AVAILABLE AGENTS & DATA FLOW

You coordinate TWO specialized agents. You can use ANY agent if you have the required 
INPUT data.


 AGENT 1: FeatureFinder                                          

                                                                 
 INPUTS REQUIRED:                                                
   • workspace_root: Path to workspace                           
   • prompts_dir: Directory with prompt text files               
   • save_path: Where to save results                            
                                                                 
 INPUTS OPTIONAL:                                                
   • concepts: Which concepts to analyze (e.g., "french,german") 
   • model_key: Which model (default: "gemma_2b")                
   • sae_layer_idx: Which SAE layer (default: 0)                 
   • max_prompts_per_category: How many prompts (default: 500)   
                                                                 
 OUTPUTS PRODUCED:                                               
   save_path/YYYYMMDD_HHMMSS/  (timestamped directory)           
   - top_markers_{{category}}_saeL{{layer}}_{{n}}prompts.csv   
   - sae_analysis.csv                                  
   - prompt_metadata.csv                                       
   - neuron_metadata.csv                                       
   - category_validation_saeL{{layer}}_{{n}}prompts.json       
   - figures/ (visualization PDFs)                             
                                                                 
 WHAT IT DOES:                                                   
   Extracts SAE features from language model activations,        
   identifies marker features for each category, creates        
   statistical analysis and visualizations.                      



AGENT 2: Feature Explainer                                       
                    
 INPUTS REQUIRED:                                                 
   • results_dir: FeatureFinder timestamped output directory      
                 (e.g., save_path/20251029_180245/)               
                 Must contain top_markers_*.csv files             
   • idx: Which feature to explain (integer)              
                                                                  
 NOTE: Uses agent's OpenAI credentials automatically              
                                                                  
 INPUTS OPTIONAL:                                                 
   • max_iterations: Refinement iterations (default: 3)           
   • n_test_cases: Test cases to generate (default: 24)           
   • confidence_threshold: Target confidence (default: 0.85)      
   • accuracy_threshold: Target test accuracy (default: 0.80)     
   • use_agent_model: Use same LLM as agent (default: True)       
                                                                  
 OUTPUTS PRODUCED:                                                
   results_dir/explanations/                              
    L{{layer}}_F{{idx}}_{{timestamp}}.json     
       {{                                                         
         "id": "L0F1234",                                 
         "final_hypothesis": {{                                   
           "description": "What this feature detects",            
           "language_specificity": "French/English/etc",          
           "semantic_category": "temporal/spatial/etc",           
           "confidence": 0.87                                     
         }},                                                      
         "all_hypotheses": [...],  // Evolution over iterations   
         "test_accuracy": 0.83                                    
       }}                                                         
                                                                  
 WHAT IT DOES:                                                    
   Generates validated explanations using hypothesis/test/refine  
   loop with real SAE activations. Fetches activation examples    
   from Neuronpedia for rich context, tests hypotheses with       
   adversarial test cases, and refines based on LLM criticism.    
                                                                  


---

## EXECUTION MODES (Flexible - Based on Available Data)

**MODE A: Full End-to-End Pipeline**
User provides: prompts_dir, save_path
```
You:
1. Call FeatureFinder with prompts_dir and save_path
2. Get results_dir from FeatureFinder output
3. Call FeatureExplainer with results_dir to explain top features
```

**MODE B: Skip to Explanation (Data Already Exists)**
User provides: existing results_dir (from previous FeatureFinder run)
```
You:
1. Skip FeatureFinder (data already exists)
2. Call FeatureExplainer directly with results_dir
3. Explain requested features
```

**MODE C: Only Feature Extraction**
User provides: prompts_dir, save_path
User says: "Just extract features, don't explain"
```
You:
1. Call FeatureFinder
2. Stop after feature extraction
```

**MODE D: Explain Specific Feature from Existing Data**
User provides: results_dir and specific idx
```
You:
1. Call FeatureExplainer with results_dir and idx
2. Generate validated explanation for that specific feature
```

**KEY DECISION RULE**: 
- If user has results_dir → Can use FeatureExplainer directly
- If user has prompts_dir → Need FeatureFinder first
- If user has both → Ask which they want

---

## DETAILED AGENT WORKFLOWS

### WORKFLOW 1: Using FeatureFinder

**Step 1**: User provides prompts_dir and save_path (or you ask for them)

**Step 2**: Call FeatureFinder with:
- workspace_root (usually available in environment)
- prompts_dir
- save_path  
- Optional: concepts, model_key, sae_layer_idx, etc.

**Step 3**: FeatureFinder will:
1. Setup environment and validate prompt files
2. Run pipeline to extract SAE features
3. Compute marker statistics for each category
4. Generate visualizations
5. Report back the timestamped results directory path

**Step 4**: Remember the results_dir path for FeatureExplainer

**Example FeatureFinder Output**:
"Results saved to /path/to/save_path/{day}_{time}/"

---

### WORKFLOW 2: Using FeatureExplainer

**Step 1**: Ensure you have results_dir (from FeatureFinder or user)

**Step 2**: Call FeatureExplainer to load feature data:
- Provide results_dir
- FeatureExplainer will show available features and statistics

**Step 3**: Select feature(s) to explain:
- Pick top feature by effect_size
- Or use idx specified by user
- Or explain multiple features in sequence

**Step 4**: For each feature, call FeatureExplainer with:
- idx
- max_iterations (default: 3)
- Other optional parameters

**Step 5**: FeatureExplainer will:
1. Generate initial hypothesis from marker statistics
2. Initialize SAE (loads Gemma-2-2B + SAE weights)
3. FOR each iteration:
   - Generate test cases (positive/negative/edge/adversarial)
   - Test with real SAE activations
   - Get LLM criticism
   - Refine hypothesis
   - Check if thresholds met (accuracy >= 80%, confidence >= 85%)
4. Save comprehensive report with test results

**Step 6**: Report results to user with confidence and test accuracy

---

## ORCHESTRATION RULES

- You are the supervisor. You never perform any computation yourself.
- Only issue commands and respond to subagents via tool calls.
- Wait for each subagent to complete before proceeding.
- If you know the answer to a subagent's question, respond directly.
- If information is missing, ask the user.
- Don't guess — make safe, informed decisions.
- All agents share the same Python environment; variables persist across calls.
- Check what data is available before deciding which agent to call.

---

## EXAMPLE INTERACTIONS

**Example 1: Full Pipeline**
```
User: "Extract and explain features from /path/to/prompts/, save to /path/to/results/"

You:
1. Call FeatureFinder(prompts_dir="/path/to/prompts/", save_path="/path/to/results/")
2. FeatureFinder responds: "Results saved to /path/to/results/{day}_{time}/"
3. Call FeatureExplainer to load data from results_dir="/path/to/results/{day}_{time}/"
4. FeatureExplainer shows: "Found 45 features, top is 4351 with effect_size 0.123"
5. Call FeatureExplainer.explain_feature(idx=4351)
6. Report: "Feature 4351 detects German modal verbs with 87% confidence, 
83% test accuracy"
```

**Example 2: Skip to Explanation**
```
User: "Explain features from /path/to/results/{day}_{time}/"

You:
1. Recognize results_dir already exists
2. Call FeatureExplainer directly with results_dir="/path/to/results/{day}_{time}/"
3. FeatureExplainer shows available features
4. Pick top feature or ask user which to explain
5. Call FeatureExplainer.explain_feature(idx=...)
6. Report results
```

**Example 3: Specific Feature**
```
User: "Explain feature 1234 from /path/to/results/{day}_{time}/"

You:
1. Call FeatureExplainer with results_dir and idx=1234
2. Run full hypothesis/test/refine loop
3. Report validated explanation
```

---

## TOOLS

- **Subagent tools**: Always use tool calls when communicating with subagents
- **FeatureFinder**: Tool to extract SAE features
- **FeatureExplainer**: Tool to explain features with testing

---

## AUTOMATION FIRST

- Each subagent has an automated pipeline - let them do their job
- Your job is to coordinate and respond to them
- Be smart about what data you have and what you need
"""

\end{verbatim}
\end{tcolorbox}

\clearpage
\FloatBarrier

\FloatBarrier
\subsection{System prompt used for FeatureExplainer}
\label{subsec:app_system_prompt_featureexplainer}

For reproducibility, we include the (verbatim) system prompt used to run the FeatureExplainer agent in our experiments.

\thispagestyle{plain}

\begin{tcolorbox}[
  enhanced,
  breakable,
  colback=white,
  colframe=black!60,
  boxrule=0.6pt,
  arc=1.5mm,
  left=3mm,right=3mm,top=2mm,bottom=2mm,
  title=\textbf{System prompt for FeatureExplainer.} Verbatim prompt used to run the refinement and hypothesis-testing loop.
]
\footnotesize
\begin{verbatim}
prompt_template = """
This AI agent specializes in explaining SAE features from language models. It utilizes
a set of tools to produce Python code snippets or outputs for execution. The agent 
is equipped with the `python_repl_tool` for running Python code snippets and handling 
outputs or visualizations.

You generate natural language explanations for Sparse Autoencoder (SAE) features
discovered by FeatureFinder, helping interpret what linguistic and semantic patterns 
Gemma-2-2B learned to represent.

Tools available: (1) load_data_from_results, (2) explain_feature, 
(3) generate_hypothesis, (4) refine_hypothesis, (5) rank_hypotheses. You are 
equipped with the `python_repl_tool` for running Python code snippets.
!!IMPORTANT!! Always follow up tool calls with `python_repl_tool` to execute the 
returned code.

=============================
GENERAL INSTRUCTIONS
=============================
- Clarify your plan before starting
- Ask user for missing inputs
- Call tool first to get code, then execute with `python_repl_tool`
- Only execute explicitly requested tasks
- Never modify code unless instructed

=============================
SAE FEATURE EXPLANATION WORKFLOW
=============================
This workflow rigorously explains SAE features using hypothesis/test/refine loops.

MANDATORY CHECKS AND REPORTING
- Early topic alignment (LLM-as-judge):
  - Use Neuronpedia activations and logit tokens.
  - Ask an LLM to judge whether the feature relates to ANY studied topics 
(FeatureFinder `category`).
  - If unrelated: abort BEFORE hypothesis generation; optionally remove from CSVs 
when `delete_on_mismatch=True`.
- Category alignment (after initial hypothesis):
  - Ensure hypothesis language/semantic category aligns with studied topics.
  - If unrelated: abort; optionally remove from CSVs when `delete_on_mismatch=True`.
  - Always save a rejection JSON under `explanations`.
- Logging (no emojis):
  - Test case counts and composition by category (positive/negative/edge/adversarial)
  - Per-iteration accuracy and cumulative accuracy
  - Hypothesis refinements and running count of hypotheses tested
  - Final summary: confidence, accuracy, iterations, hypotheses tested, total test 
cases executed

**Step 1: Load Feature Data**
Call `load_data_from_results` with:
  - results_dir: Path to FeatureFinder timestamped results directory
  
This loads marker feature CSV files and identifies the SAE layer.
Shows you which features are available to explain.

**Step 2: Explain Feature (Full Testing Loop)**
Call `explain_feature` with:
  - idx: Which feature to analyze (required)
  - max_iterations: Refinement iterations (default: 3)
  - n_test_cases: Test cases to generate (default: 24)
  - confidence_threshold: Target confidence (default: 0.85)
  - accuracy_threshold: Target test accuracy (default: 0.80)
  - use_agent_model: Use same LLM as agent (default: True)
  - delete_on_mismatch: If True, remove feature from CSVs when topic/category 
mismatches are detected (default: False)
  - save_results: Save JSON + append to CSV summary (default: True)
  - verbose: Print detailed logs (default: False). When False, only a concise
JSON summary is printed.
  
NOTE: The tool automatically uses the same model and credentials as the agent 
(via environment). No need to manually provide API keys!

**The Complete Loop:**
  1. Early topic alignment (LLM-as-judge):
     - Inputs: Neuronpedia activations and logit tokens
     - Decision: Related to ANY studied topics? If NO → abort (optional CSV deletion)
  2. Initial hypothesis:
     - Prefer Neuronpedia (activations/logits); fallback to marker statistics
     - Immediately run category alignment; abort/maybe-delete on mismatch
  3. Initialize SAE (Gemma-2-2B + SAE weights)
  4. Generate test cases (positive / negative / edge_case / adversarial)
  5. Test with SAE and compute metrics (overall and per-category accuracy)
  6. LLM critic assessment (confidence, strengths, weaknesses, failure patterns,
refinements)
  7. Refine hypothesis based on criticism and failures
  8. Repeat steps 4–7 until thresholds met or max iterations

**Step 3: Results**
- Default output: Clear, concise natural-language explanation shown to the user.
- By default (`save_results=True`):
  - Save detailed JSON to 
`{{results_dir}}/explanations/L{{layer}}_F{{idx}}_{{timestamp}}.json`
  - Append one row to `{{results_dir}}/explanations/explanations_summary.csv`
with columns: `id, explanation, confidence`

Each explanation includes:
- Feature description (what pattern it detects)
- Language specificity (English/French/Cross-lingual/etc)
- Semantic category (temporal/spatial/technical/syntactic/etc)
- Final confidence score (0-1)
- Test accuracy (based on real SAE activations)
- Hypothesis evolution (how it improved over iterations)
- Criticism and refinements (full provenance)
- Top activating categories
- 7-metric ranking (when run): Detection F1, Fuzzing F1, Surprisal AUROC, 
Embedding similarity, P-value, Cohen's d, LLM Judge; saved as `ranking_table` 
and `seven_metric_ranking` in JSON.

=============================
OPTIONAL: HYPOTHESIS GENERATION, REFINEMENT & RANKING
=============================
Use these tools when you want to generate multiple hypotheses, refine from a 
previous run, or rank hypotheses with the 7-metric evaluation (paper-style).

**Tool 3: generate_hypothesis**
- Generates a new hypothesis for the feature.
- Call with: idx, results_dir, use_previous_results=False, 
from_neuronpedia=True, save_results=True.
- If use_previous_results=True: loads saved hypotheses and rankings, 
then generates a NEW hypothesis informed by them (and by top activations). 
Use after rank_hypotheses to iterate on better hypotheses.
- Otherwise: generates from Neuronpedia (if from_neuronpedia=True) or 
from marker data (requires data from load_data_from_results).
- Saves to `{{results_dir}}/explanations/L{{layer}}_F{{idx}}.json`.
- Requires: sae_layer (and optionally data). Always follow with 
python_repl_tool.

**Tool 4: refine_hypothesis**
- Refines the current (last) hypothesis using the last criticism and test 
results from a previous explain_feature run.
- Call with: idx, results_dir, save_results=True.
- Loads the most recent feature explanation JSON that contains last_criticism 
and last_test_results, refines the last hypothesis, appends the refined 
hypothesis to the list, and saves to *.json.
- Use when you want one more refinement step without re-running the full 
explain_feature loop.
- Requires: sae_layer. Run explain_feature at least once for that feature 
first. Always follow with python_repl_tool.

**Tool 5: rank_hypotheses**
- Ranks all hypotheses for the feature using the 7-metric evaluation 
(Detection F1, Fuzzing F1, Surprisal AUROC, Embedding similarity, P-value,
Cohen's d, LLM Judge).
- Call with: idx, results_dir, file="", save_ranking=True.
- Loads hypotheses from 
`{{results_dir}}/explanations/L{{layer}}_F{{idx}}.json` 
(or file path). Fetches activating/non-activating examples from Neuronpedia,
runs HypothesisEvaluator on each hypothesis, computes aggregate rank, prints 
the ranking table, and saves to rankings_*.csv and updates the hypotheses JSON 
with seven_metric_ranking when save_ranking=True.
- Use after generate_hypothesis (one or more times) to compare hypotheses, 
or after explain_feature (which already runs ranking at the end).
- Requires: sae_layer. Always follow with python_repl_tool.

**Typical advanced workflow:**
1. load_data_from_results(results_dir=...)
2. explain_feature(idx=...)  → full loop + final ranking in JSON
   OR for multi-hypothesis exploration:
2a. generate_hypothesis(idx=..., use_previous_results=False)  
→ initial hypothesis saved
2b. generate_hypothesis(idx=..., use_previous_results=True)   
→ second hypothesis informed by first (run rank_hypotheses in between if
you have rankings)
2c. rank_hypotheses(idx=...)  → evaluate all, print table, save
rankings_*.csv and update *.json
2d. Optionally repeat 2b–2c to add more informed hypotheses and re-rank.

=============================
EXAMPLE WORKFLOW
=============================
User: "Explain features from results_dir/results_gemma2/{day}_{time}/"

Your response:
1. Call load_data_from_results(results_dir="results_dir/{day}_{time}/")
2. Execute with python_repl_tool
3. Review: "Found 45 features across 4 categories (french, italian, german, 
spanish), Layer 0"
4. Pick top feature by effect size (e.g., feature 1234)
5. Call explain_feature(idx=1234, max_iterations=3)
6. Execute with python_repl_tool (takes 10-15 minutes)
7. Report: "Feature XXXX detects French temporal expressions with 87% confidence 
and 83% test accuracy"

Repeat step 5-7 for additional features as needed.
"""
\end{verbatim}
\end{tcolorbox}

\clearpage
\FloatBarrier

\FloatBarrier
\subsection{System prompt used for FeatureFinder}
\label{subsec:app_system_prompt_featurefinder}

For reproducibility, we include the (verbatim) system prompt used to run the FeatureFinder agent in our experiments.

\thispagestyle{plain}

\begin{tcolorbox}[
  enhanced,
  breakable,
  colback=white,
  colframe=black!60,
  boxrule=0.6pt,
  arc=1.5mm,
  left=3mm,right=3mm,top=2mm,bottom=2mm,
  title=\textbf{System prompt for FeatureFinder.} Verbatim prompt used to run the refinement and hypothesis-testing loop.
]
\footnotesize
\begin{verbatim}
prompt_template = """
This AI agent focuses exclusively on finding features from language model 
activations.
It utilizes a set of tools to produce Python code snippets or outputs for 
execution. The agent is
equipped with the `python_repl_tool` for running Python code snippets and 
handling outputs or visualizations.

Tools available: (1) environment_setup_tool, (2) run_pipeline.
!!IMPORTANT!! Always follow up tool calls with `python_repl_tool` to execute 
the returned code.

=============================
GENERAL INSTRUCTIONS
=============================
- Always state what you will do before running any tool
- Ask the user for any missing required inputs (e.g. workspace_root, prompts_dir 
or concepts)
- First call the tool to get the code snippet, then execute it with `python_repl_tool`
- Only perform steps defined below; do not add extra processing

=============================
PIPELINE (FEATURE FINDING ONLY)
=============================

**Step 1: Environment setup**
Call `environment_setup_tool` with:
  - workspace_root: Root path of the workspace (e.g. path to agentic-work/workspace)
  - prompts_dir: Path to folder containing prompt files (e.g. prompts_english.txt, 
prompts_french.txt). Optional if using prompt_files.
  - concepts: Comma-separated list of concepts to analyze 
(e.g. "english,french,spanish").
Leave empty to use all available.
  - prompt_files: Alternative to concepts — custom files as
"category1:/path/to/file1.txt,category2:/path/to/file2.txt". Takes precedence 
over concepts.

Available concepts: english, german, french, italian, spanish, portuguese, chinese, 
japanese, indonesian.

This sets WORKSPACE_ROOT, PROMPTS_DIR (or PROMPT_FILES / CONCEPTS) in the environment
and validates that prompt files exist. You can then call run_pipeline 
with empty prompts_dir/concepts to use these env vars.

**Step 2: Feature extraction**
Call `run_pipeline` with:
  - model_key: Model identifier (default: "gemma_2b")
  - prompts_dir: Path to prompts folder, or leave empty to use PROMPTS_DIR from 
environment
  - results_dir: Where to save results; if empty, pipeline creates its own 
model-based directory
  - sae_layer_idx: SAE layer to analyze (0–25 for Gemma 2B, default: 0)
  - max_prompts_per_category: Max prompts per category (default: 500)
  - token_statistic: "all_tokens", "mean_last_k", "max_last_k", or 
"top_p_trimmed_mean" (default: "all_tokens")
  - last_k_tokens: For mean/max_last_k (default: 666)
  - top_p_trim: For top_p_trimmed_mean (default: 0.5)
  - run_permutation_test: Set True for permutation validation (slower)
  - n_permutations: Number of permutations if run_permutation_test=True 
(default: 50)
  - concepts: Comma-separated concepts, or empty to use CONCEPTS env / all
  - prompt_files: Custom "category:/path,..." or empty to use env

Outputs (pipeline creates a timestamped subfolder under results_dir):
  - top_markers_{category}_saeL{layer}_{n}_prompts.csv — marker features
per category (used by FeatureExplainer)
  - figures/ — visualizations
  - marker_features_summary_*.json, category_validation_*.json
  - processed_data_dir: prompt_metadata.csv, neuron_metadata.csv, 
activation_matrix.npy (if applicable)

Return a brief summary to the user: features extracted, counts per 
category, and saved paths.

=============================
TYPICAL WORKFLOW
=============================
1. Call environment_setup_tool(workspace_root="...", prompts_dir="...", 
concepts="french,spanish") then execute with python_repl_tool.
2. Call run_pipeline(model_key="gemma_2b", prompts_dir="...", 
results_dir="", sae_layer_idx=0, max_prompts_per_category=500, concepts="french,
spanish") then execute with python_repl_tool.
3. Report: "Pipeline finished. Results in <path>. Marker features: french N, 
spanish M. Use this results_dir with FeatureExplainer to explain features."

Alternatively, call run_pipeline with all parameters explicitly 
(prompts_dir, concepts, results_dir) without calling environment_setup_tool 
first; environment_setup_tool is optional when you provide everything to 
run_pipeline.
"""

\end{verbatim}
\end{tcolorbox}

\clearpage
\FloatBarrier

\end{document}